\documentclass[runningheads]{llncs}

\usepackage{eccv}

\usepackage{eccvabbrv}

\usepackage{graphicx}
\usepackage{booktabs}
\usepackage[table]{xcolor}
\usepackage{multirow}
\definecolor{oursbg}{RGB}{255, 248, 190}   

\usepackage[accsupp]{axessibility}  

\usepackage{hyperref}

\usepackage{orcidlink}

\begin{document}

\title{PRISM: Latent Composition Consistency for \\ Single-Image Reflection Removal}

\titlerunning{PRISM: Latent Composition Consistency for Single-Image Reflection Removal}

\author{Junseong Shin\inst{1} \and
Tae Hyun Kim\inst{1, 2}\thanks{Corresponding author}}

\authorrunning{J.~Shin and T.H.~Kim}

\institute{Department of Artificial Intelligence, Hanyang University, Seoul, South Korea \and
Department of Computer Science, Hanyang University, Seoul, South Korea\\
\email{\{junsung6140, taehyunkim\}@hanyang.ac.kr}}

\maketitle

\begin{abstract}
Single-image reflection removal (SIRR) seeks to recover the transmission layer from a mixture corrupted by reflections---a severely ill-posed problem. 
Existing methods operate in pixel space, where the nonlinear sRGB formation model entangles the two layers and limits generalization.
We observe that pretrained VAE latent spaces exhibit substantially lower coherence between image layers compared to pixel space, providing a more favorable working space for decomposition.
Building on this finding, we propose \textbf{PRISM} (Pretrained-latent Reflection Image Separation Model), which reinterprets SIRR as a latent linear separation problem. Under an approximate additive formulation in latent space, PRISM learns a flow matching velocity field on a pretrained FLUX backbone that recovers both transmission and reflection in a single forward pass. To enforce robust disentanglement, we introduce a Latent Composition Consistency (LCC) strategy that constructs synthetic mixtures by swapping reflection latents across samples and enforces consistent decomposition via a cycle loss. We further propose a Layer Contrastive Separation (LCS) loss that promotes semantic separation between layers through patch-level contrastive learning, without requiring explicit reflection targets. Experiments on six benchmarks demonstrate that PRISM consistently outperforms state-of-the-art methods by significant margins, with strong generalization to in-the-wild images. Our project page is available at \url{https://junsung6140.github.io/prism/}.
\keywords{Reflection Removal \and Flow Matching \and Contrastive Learning}
\end{abstract}

\section{Introduction}

Photographs taken through glass windows or other transparent surfaces routinely capture unwanted reflections superimposed on the scene of interest.
Beyond degrading visual quality, such reflections harm downstream tasks including 3D Gaussian Splatting~\cite{wang2024dc}, face recognition~\cite{wan2021face}, and depth estimation~\cite{chang2020joint}.
Single-image reflection removal (SIRR) seeks to recover the transmission layer $T$ from an observed mixture $I = T + R$, where $R$ is the reflection component~\cite{Nayar1997}.
This problem is severely ill-posed, as infinitely many $(T, R)$ pairs satisfy the equation for a given $I$.

\begin{figure}[t]
    \centering
    \includegraphics[width=\linewidth]{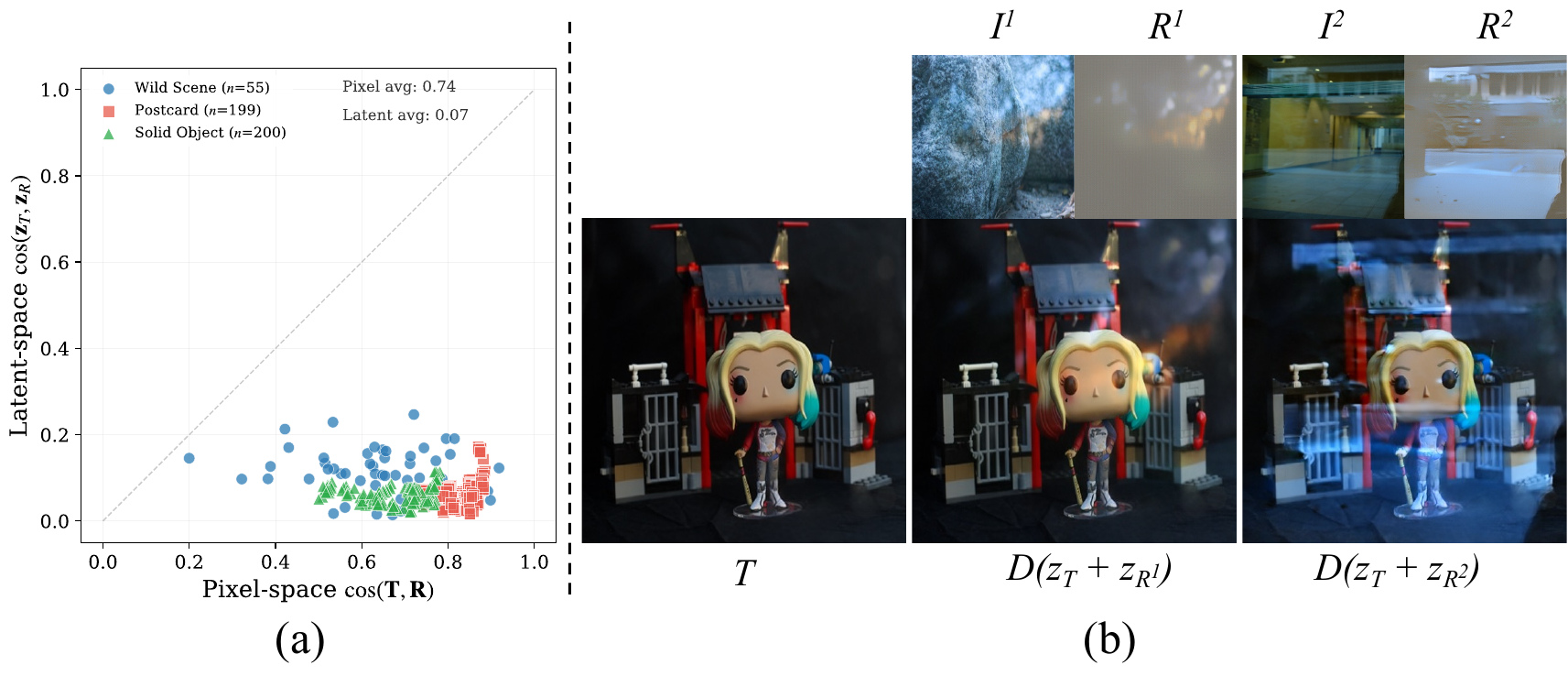}
    \caption{\textbf{(a)} Cosine similarity between transmission and
    reflection in pixel space vs.\ FLUX VAE~\cite{blackforest2024flux} latent space on 454 pairs from
    SIR$^2$~\cite{wan2017benchmarking}. All points lie below the diagonal,
    confirming consistently lower coherence in latent space.
    \textbf{(b)} Latent swap results: composing $\hat{z}_T$ with reflections
    $\hat{z}_R^1$, $\hat{z}_R^2$ separated from different images and decoding
    via the VAE decoder $\mathcal{D}$ yields realistic mixtures that preserve transmission
    content while swapping only the reflection, demonstrating well-disentangled
    decomposition.}
    \label{fig:intro}
\end{figure}

Prior works address this ambiguity through hand-crafted priors~\cite{kim2020single,li2014single,shih2015reflection}, learnable residue terms~\cite{hu2023single,zheng2021single}, or data-driven architectures~\cite{zhao2025reversible,hu2024single,hu2025dereflection}.
However, operating in pixel space imposes two fundamental limitations: the additive model is only a coarse approximation under nonlinear camera responses, and architectures must simultaneously capture low-level texture and high-level semantics, often resulting in entangled features.
While the RAW sensor domain offers a more faithful additive model~\cite{kee2025removing}, it requires specialized hardware inapplicable to existing sRGB imagery.

\textbf{Why latent space?} We pursue an orthogonal direction: lifting the decomposition into the latent space of a pretrained generative model~\cite{kingma2013auto,blackforest2024flux,rombach2022high}.
As shown in Fig.~\ref{fig:intro}(a), we measure the cosine similarity between transmission and reflection across 454 real image pairs from SIR$^2$~\cite{wan2017benchmarking}.
In pixel space, $T$ and $R$ are highly entangled (avg.\ $\cos = 0.74$), whereas in the FLUX VAE latent space the cosine similarity drops to $0.07$ — a reduction shared by any pair of encoded images, indicating broad decorrelation rather than a layer-specific property.
This substantially lower coherence suggests that the latent space serves as a more favorable working space for decomposition.
Leveraging this property, we adopt an approximate additive formulation $z_I \approx z_T + z_R$, where $z = \mathcal{E}(\cdot)$ denotes the output of a pretrained VAE encoder $\mathcal{E}$ and $\mathcal{D}(\cdot)$ its decoder.
We model the transport $z_I \to z_T$ via a flow matching velocity field $v_\theta$.
The velocity target satisfies $v^{*} \approx -z_R$, enabling both components to be recovered in a single forward pass without dual-branch architectures.

We term our method \textbf{PRISM} (\textbf{P}retrained-latent \textbf{R}eflection \textbf{I}mage \textbf{S}eparation \textbf{M}odel).
A key challenge is preventing degenerate solutions where the model entangles scene-specific information across components.
The low mutual coherence of latent representations offers a unique opportunity here: since the overlap between layers is substantially lower in latent space, composing the transmission of one image with the reflection of another yields a semantically valid mixture (Fig.~\ref{fig:intro}(b)) — something far less effective in pixel space where the two layers overlap heavily.
We exploit this property with a \textbf{Latent Composition Consistency (LCC)} strategy that swaps reflection components across images to construct synthetic latent compositions, enforcing that the model decomposes them consistently.

Since the additive formulation $z_I \approx z_T + z_R$ is only approximate, an explicit reflection regression target would require encoding a pixel-space residual ($\mathcal{E}(I-T)$), propagating nonlinearity errors into the training signal.
We therefore introduce a \textbf{Layer Contrastive Separation (LCS)} loss that enforces semantic separation between transmission and reflection without requiring any explicit reflection target.

Our contributions are as follows:
\begin{itemize}
  \item We provide empirical evidence that pretrained VAE latent spaces exhibit significantly lower coherence between image layers compared to pixel space, and show that this property enables effective latent-space decomposition.

  \item We propose \textbf{PRISM}, a latent flow matching framework that reinterprets SIRR as a latent linear separation problem, recovering both transmission and reflection in a single forward pass without dual-branch architectures.

  \item We introduce a \textbf{Latent Composition Consistency (LCC)} training strategy that leverages the low coherence of latent space to construct semantically valid cross-image compositions, enforcing decomposition robustness across arbitrary latent mixtures.

  \item We introduce a \textbf{Layer Contrastive Separation (LCS)} loss that enforces semantic separation between transmission and reflection without requiring exact additive reconstruction, directly addressing the non-exact nature of the latent additive assumption.
\end{itemize}

\section{Related Works}

\subsection{Single-Image Reflection Removal}

Early work models the observed image as $I = T + R$, where $T$ and $R$ denote the transmission and reflection layers~\cite{Nayar1997}.
Subsequent refinements incorporate ghosting cues~\cite{shih2015reflection}, alpha-matting coefficients~\cite{li2014single}, channel-wise blending scalars~\cite{kim2020single}, and a learnable residue term $\Phi(T,R)$ to capture nonlinear interactions~\cite{hu2023single}.
Operating in the RAW sensor domain makes this additive model more physically faithful~\cite{kee2025removing}, but requires specialized hardware unavailable for standard sRGB imagery.

With paired datasets, supervised networks have become dominant.
Single-stream methods such as CEILNet~\cite{fan2017generic}, Zhang \textit{et al.}~\cite{zhang2018single}, and ERRNet~\cite{wei2019single} learn direct mappings to the transmission layer via task-specific architectures and losses.
Recognizing the complementary nature of the two layers, dual-stream architectures model both jointly: IBCLN~\cite{li2020single} uses a recurrent dual-branch LSTM; YTMT~\cite{hu2021trash} introduces a symmetric ReLU-based interaction strategy; LASIRR~\cite{dong2021location} adds explicit reflection localization with confidence maps; and \cite{zheng2021single}, DSRNet~\cite{hu2023single}, and DSIT~\cite{hu2024single} progressively advance feature interaction quality via gated modules, learnable residue terms, and dual-attention mechanisms, respectively.
RDNet~\cite{zhao2025reversible} further addresses information loss in dual-stream interactions through multi-level reversible encoders, while Zhu \textit{et al.}~\cite{zhu2024revisiting} contribute a real-world dataset to mitigate data scarcity.
More recently, diffusion-based priors~\cite{ho2020denoising} have been brought to bear on SIRR: DAI~\cite{hu2025dereflection} demonstrates that constraining predictions to a natural-image manifold substantially improves real-world generalization, though it requires fine-tuning the Stable Diffusion~\cite{rombach2022high} VAE to compensate for its limited reconstruction fidelity.
\textit{Despite this progress, all of the above methods operate in pixel space and are susceptible to the nonlinear sRGB formation model; none enforces a structured additive decomposition in the representation space of a generative model.}

\subsection{Flow Matching and Latent Generative Models}

VAEs~\cite{kingma2013auto} and latent diffusion models~\cite{rombach2022high} establish that compressed latent spaces are smooth, semantically structured, and amenable to compositional arithmetic—properties exploited for flexible image editing~\cite{meng2021sdedit} and generation.
Flow matching~\cite{lipman2022flow} learns a velocity field transporting samples along straight trajectories; Rectified Flow~\cite{liu2022flow} further straightens these paths, and consistency models~\cite{song2023consistency} push this to single-step generation.
FLUX~\cite{blackforest2024flux} instantiates these principles at scale, providing a pretrained latent space well-suited for task-specific fine-tuning.
Generative priors have also been applied to image restoration~\cite{ho2020denoising}, with diffusion- and flow-based methods~\cite{luo2023image,xia2023diffir,lin2024diffbir,zhu2024flowie} showing that constraining predictions to a natural-image manifold substantially improves real-world generalization.
\textit{Unlike these methods, which repurpose generative models as pixel-space priors or require iterative sampling, we directly exploit the additive structure of the latent space for single-pass reflection decomposition.}

\subsection{Contrastive Learning for Image Restoration}

Contrastive learning~\cite{oord2018representation,chen2020simple,he2020momentum} trains representations by pulling positive pairs together while pushing negatives apart, and has been increasingly adopted in low-level vision tasks.
AECR-Net~\cite{wu2021contrastive} first introduced contrastive regularization for image dehazing by treating hazy images as negatives and their clean counterparts as positives, establishing the paradigm of using degradation-clean pairs as contrastive supervision.
CUT~\cite{park2020contrastive} further demonstrated that \textit{patch-wise} contrastive objectives can effectively align local structures in unpaired image-to-image translation, inspiring subsequent patch-level formulations.
More recently, task-agnostic contrastive strategies~\cite{wu2024learning} have been proposed to improve restoration quality across diverse degradation types.
\textit{Unlike these works, which apply contrastive objectives to improve single-target reconstruction quality, our Layer Contrastive Separation loss repurposes this framework for cross-layer disentanglement between transmission and reflection.}

\section{Method}

\subsection{Preliminary: Latent Flow Matching and FLUX}

\subsubsection{Flow Matching.}
Flow matching~\cite{lipman2022flow} learns a velocity field $v_\theta$ that
transports samples from a source distribution $p_0$ to a target distribution
$p_1$ along straight trajectories.
Given a sample pair $(x_0, x_1)$, the conditional flow is defined as the linear
interpolation $x_t = (1-t)x_0 + t x_1$ for $t \in [0, 1]$, with constant
velocity target $v^* = x_1 - x_0$.
At inference, the one-step estimate recovers the target as
$x_1 = x_0 + v_\theta(x_0, 0)$.

\subsubsection{Latent Flow Matching.}
Rather than operating in pixel space, latent flow matching~\cite{esser2024scaling,wu2025qwenimage} performs this
transport in the compressed representation of a pretrained
VAE~\cite{kingma2013auto}.
An encoder $\mathcal{E}$ maps an image $I$ to a latent
$z = \mathcal{E}(I) \in \mathbb{R}^{16 \times H/8 \times W/8}$
(16 channels, $8\times$ spatial downsampling) and a decoder $\mathcal{D}$
inverts this mapping.
The latent space is semantically structured and exhibits smooth compositional
properties~\cite{rombach2022high}, making it
well-suited for the additive decomposition we exploit in Sec.~\ref{sec:formulation}.
Crucially, FLUX's VAE achieves high reconstruction fidelity without additional fine-tuning, unlike Stable Diffusion-based approaches~\cite{rombach2022high,hu2025dereflection} that require VAE fine-tuning to compensate for reconstruction artifacts.

\subsubsection{FLUX.}
FLUX~\cite{blackforest2024flux} is a large-scale instantiation of latent flow
matching built on a Multi-Modal Diffusion Transformer (MM-DiT)
backbone~\cite{esser2024scaling}.
We adopt FLUX as the backbone for our reflection removal framework,
as detailed in Sec.~\ref{sec:latent_fm}.
Conventional diffusion-based image-to-image methods begin from a noisy latent
obtained by adding noise to the encoded input, and iteratively denoise toward
the target.
FLUX, as a flow matching model, supports starting directly from a clean input
latent without any noise injection, since the straight-line trajectory allows
recovery at any timestep.
We exploit this property: PRISM takes the clean input latent
$z_I = \mathcal{E}(I)$ as the starting point and recovers the transmission
latent in a single forward pass.

\subsection{Problem Formulation}
\label{sec:formulation}

We adopt the standard additive image formation model~\cite{Nayar1997,zhang2018single,wan2017benchmarking}, where the observed image
$I$ is expressed as the superposition of a transmission layer $T$ and a
reflection layer $R$:
\begin{equation}
I = T + R.
\end{equation}
This formulation serves only as an approximation under real imaging conditions
due to nonlinear camera responses, blur, and attenuation effects.
Instead of refining the physical model in pixel space, we perform decomposition in the VAE latent space introduced above, where encoded representations exhibit substantially lower mutual coherence (Fig.~\ref{fig:intro}(a)).
We adopt an approximate additive formulation:
\begin{equation}
z_I \approx z_T + z_R,
\end{equation}
where $z_I = \mathcal{E}(I)$, $z_T = \mathcal{E}(T)$, and
$z_R = \mathcal{E}(R)$.
Although this constraint does not hold exactly due to encoder nonlinearity, the low coherence in latent space eases the regression and makes it a practically effective working assumption, casting SIRR as a \emph{latent linear separation} problem.

\begin{figure}[t]
    \centering
    \includegraphics[width=0.99\linewidth]{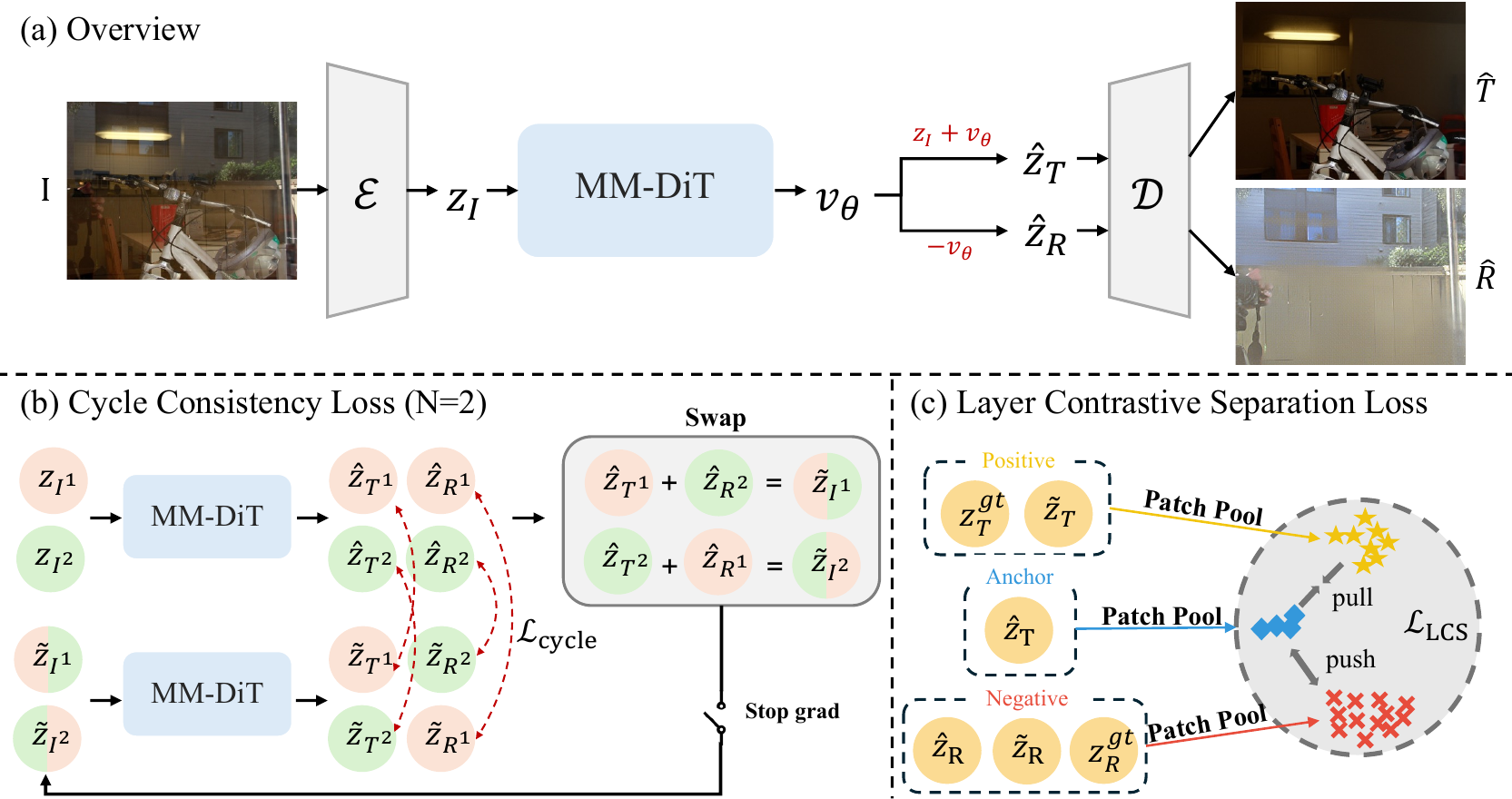}
    \caption{\textbf{Overview of PRISM.}
    \textbf{(a)} MM-DiT predicts $v_\theta$ from $z_I = \mathcal{E}(I)$
    (VAE encoder $\mathcal{E}$, decoder $\mathcal{D}$),
    yielding $\hat{z}_T = z_I + v_\theta$ and $\hat{z}_R = -v_\theta$ in a single pass;
    both are decoded by $\mathcal{D}$ to produce $\hat{T}$ and $\hat{R}$.
    \textbf{(b)} Reflection latents are cyclically swapped across batch samples
    to form synthetic mixtures $\tilde{z}_I$; a second forward pass enforces
    consistent decomposition ($\mathcal{L}_\text{cycle}$).
    \textbf{(c)} Patch-pooled features of the cycle-recovered transmission $\tilde{z}_T$
    are pulled toward positive (transmission) and pushed from negative (reflection) samples
    via InfoNCE ($\mathcal{L}_\text{LCS}$).}
    \label{fig:overview}
\end{figure}

\subsection{Latent Flow Matching for Reflection Removal}
\label{sec:latent_fm}

Given the latent additive formulation $z_I \approx z_T + z_R$, our goal is to
estimate $z_T$ from $z_I$ alone. We model this transformation using a flow 
matching framework, where a velocity field $v_\theta$ is learned to transport 
the input latent $z_I$ toward the transmission latent $z_T$. Specifically, 
we define the velocity target as:
\begin{equation}
v^* = z_T - z_I \approx -z_R,
\end{equation}
which reveals a key property: under the additive latent assumption,
the velocity field approximates the negative of the reflection latent,
allowing both components to be recovered in a single forward pass
without any dual-branch architecture or dedicated reflection loss
(Fig.~\ref{fig:overview}(a)):
\begin{equation}
\hat{z}_T = z_I + v_\theta(z_I, 0), \qquad \hat{z}_R = -v_\theta(z_I, 0).
\end{equation}

We implement $v_\theta$ by fine-tuning the full parameters of a pretrained
FLUX model~\cite{blackforest2024flux}, with $z_I$ serving directly as the
starting point of the flow. Rather than sampling $t \sim \mathcal{U}[0,1]$ during training, we fix $t=0$
and train with a single-step objective directly on the input latent $z_I$,
which reduces training cost and is consistent with one-step inference:
\begin{equation}
\mathcal{L}_{\text{latent}} = \| v_\theta(z_I, 0) - v^* \|^2_2.
\end{equation}
To further improve perceptual quality, we additionally supervise the decoded 
transmission prediction in pixel space using a combination of $\ell_1$ and 
LPIPS~\cite{zhang2018unreasonable} losses:
\begin{equation}
\mathcal{L}_{\text{pixel}} = \| \mathcal{D}(\hat{z}_T) - T \|_1
+ \lambda_{\text{lpips}} \cdot \mathcal{L}_{\text{LPIPS}}
(\mathcal{D}(\hat{z}_T), T),
\end{equation}

\noindent\textbf{Inference.}
At inference, we apply a single forward pass at $t=0$, so that $\hat{z}_T = z_I + v_\theta(z_I, 0)$ directly.
This one-step scheme is consistent with the linear flow matching objective~\cite{lipman2022flow}, where the ground-truth velocity field is constant along the trajectory; we train and infer at $t=0$ for simplicity and empirical stability.

\subsection{Latent Composition Consistency}

While the latent flow matching objective encourages accurate transmission
recovery, it does not explicitly prevent entanglement between $\hat{z}_T$
and $\hat{z}_R$: a degenerate solution exists where the model encodes
scene-specific information into both components without achieving true
disentanglement. We address this with two complementary losses —
$\mathcal{L}_{\text{cycle}}$ and $\mathcal{L}_{\text{LCS}}$ — both
grounded in the \textbf{Latent Composition Consistency (LCC)} strategy described below.

\subsubsection{Latent Composition Consistency (LCC).}
The low mutual coherence of transmission and reflection in latent space
(Fig.~\ref{fig:intro}(a)) enables a unique training strategy.
Since the two components exhibit substantially lower overlap, composing
the transmission of one image with the reflection of another yields a
semantically valid mixture — something far less effective in pixel space.
A well-disentangled model should therefore decompose not only real mixtures
but any such synthetic composition.
We operationalize this by cyclically swapping reflection components across
the $N$ samples of a training batch (Fig.~\ref{fig:overview}(b)):
\begin{equation}
\tilde{z}_I^i = \hat{z}_T^i + \hat{z}_R^{(i \bmod N) + 1}, \quad i \in \{1, \ldots, N\}.
\end{equation}
A second forward pass on each synthetic mixture then recovers the decomposition:
\begin{equation}
\tilde{v}^i = v_\theta(\tilde{z}_I^i, 0), \qquad
\tilde{z}_T^i = \tilde{z}_I^i + \tilde{v}^i, \qquad
\tilde{z}_R^i = -\tilde{v}^i.
\end{equation}

\subsubsection{Cycle Consistency Loss ($\mathcal{L}_{\text{cycle}}$).}
The cycle loss enforces that the second-pass decomposition recovers the
original components, averaged over the batch:
\begin{equation}
\mathcal{L}_{\text{cycle}} = \frac{1}{N}\sum_{i=1}^{N}\frac{1}{2}\left(
\| \tilde{z}_T^i - \texttt{sg}(\hat{z}_T^i) \|_2^2 +
\| \tilde{z}_R^i - \texttt{sg}\!\left(\hat{z}_R^{(i \bmod N)+1}\right) \|_2^2
\right),
\end{equation}
where $\texttt{sg}(\cdot)$ is the stop-gradient operator.
Without it, the model could trivially minimize $\mathcal{L}_{\text{cycle}}$
by collapsing all first-pass predictions to the same value (mode collapse).
Stop-gradient fixes the first-pass outputs as stable targets, so that
the second pass learns to match a consistent decomposition anchored by $\mathcal{L}_{\text{latent}}$ and $\mathcal{L}_{\text{pixel}}$.

\subsubsection{Layer Contrastive Separation Loss ($\mathcal{L}_{\text{LCS}}$).}
While $\mathcal{L}_{\text{latent}}$ supervises the velocity target $v^* = z_T - z_I$ (which only requires $\mathcal{E}(T)$ and $\mathcal{E}(I)$), it does not explicitly enforce that the resulting $\hat{z}_T$ and $\hat{z}_R$ are semantically disentangled.
Introducing an explicit reflection regression target $\mathcal{E}(I-T)$ would be problematic, since $\mathcal{E}(I-T) \neq \mathcal{E}(I) - \mathcal{E}(T)$ due to encoder nonlinearity.
Instead, we exploit the low mutual coherence of $z_T$ and $z_R$
(cosine similarity $0.07$ in latent vs.\ $0.74$ in pixel space) and enforce
semantic separation via contrastive learning, without requiring any explicit reflection target.
As illustrated in Fig.~\ref{fig:overview}(c), we build on the InfoNCE
framework~\cite{oord2018representation}, using the cycle-recovered
transmission $\tilde{z}_T$ as an anchor, with transmission latents
($\hat{z}_T$, $z_T^{\text{gt}}$) as positives and reflection latents
($\hat{z}_R$, $\tilde{z}_R$, $z_R^{\text{gt}}$) as negatives.

Since reflection and transmission exhibit spatially heterogeneous
entanglement, we extract \textbf{patch-based} representations: the latent map
is divided into non-overlapping $p \times p$ patches, each independently
pooled and $\ell_2$-normalized, yielding $K = (H_z/p)\times(W_z/p)$
contrastive pairs per sample.
Letting $f_k(\cdot)$ denote the patch-pool operation for the $k$-th patch,
and $\sigma_k(z) = f_k(\tilde{z}_T)^\top f_k(z) / \tau$ the scaled similarity to the anchor:
\begin{equation}
\mathcal{L}_{\text{LCS}} = -\mathbb{E}\left[
\frac{1}{K}\sum_{k=1}^{K}
\log \frac{
  \sum_{s \in \mathcal{P}} e^{\sigma_k(z^s)}
}{
  \sum_{s \in \mathcal{P}} e^{\sigma_k(z^s)}
  + \sum_{n \in \mathcal{N}} e^{\sigma_k(z^n)}
}
\right],
\end{equation}
where $\mathcal{P}$ and $\mathcal{N}$ are the positive and negative sets,
$\tau = 0.1$ is the temperature, and $H_z\!=\!H/8$, $W_z\!=\!W/8$.
We set $p = 4$ as the default patch size (see ablation in Sec.~4.3).
GT reflection negatives are included only when ground-truth annotations are available.

\subsection{Training Objective}

The final training objective combines all four loss terms:
\begin{equation}
\mathcal{L} = \lambda_{\text{lat}} \mathcal{L}_{\text{latent}}
+ \lambda_{\text{pix}} \mathcal{L}_{\text{pixel}}
+ \lambda_{\text{cycle}} \mathcal{L}_{\text{cycle}}
+ \lambda_{\text{LCS}} \mathcal{L}_{\text{LCS}},
\end{equation}
where $\lambda_{\text{lat}}$, $\lambda_{\text{pix}}$, 
$\lambda_{\text{cycle}}$, and $\lambda_{\text{LCS}}$ are weighting 
coefficients that balance the contribution of each term. The latent 
and pixel losses provide direct supervision for transmission recovery, 
while the cycle and Layer Contrastive Separation losses promote disentanglement without
requiring pixel-space reflection ground truth.

\begin{table*}[t]
\centering
\setlength{\tabcolsep}{5pt}
\caption{Benchmark results of various SIRR methods on Real(20), Object(200), Postcard(199), Wild(55), Nature(20), and SIR$^2$(454) datasets.}
\label{tab:benchmark_updated}
\resizebox{\textwidth}{!}{
\begin{tabular}{l|cc|cc|cc|cc|cc|cc}
\toprule
\multirow{2}{*}{Methods}
& \multicolumn{2}{c|}{\textbf{Real(20)}}
& \multicolumn{2}{c|}{\textbf{Object(200)}}
& \multicolumn{2}{c|}{\textbf{Postcard(199)}}
& \multicolumn{2}{c|}{\textbf{Wild(55)}}
& \multicolumn{2}{c|}{\textbf{Nature(20)}}
& \multicolumn{2}{c}{\textbf{SIR$^2$(454)}} \\
& PSNR & SSIM
& PSNR & SSIM
& PSNR & SSIM
& PSNR & SSIM
& PSNR & SSIM
& PSNR & SSIM \\
\midrule
ERRNet~(CVPR'19) & 22.89 & 0.803 & 24.87 & 0.896 & 22.04 & 0.876 & 24.25 & 0.853 & 20.58 & 0.756 & 23.41 & 0.874 \\
IBCLN~(CVPR'20) & 21.86 & 0.762 & 24.87 & 0.893 & 23.39 & 0.875 & 24.71 & 0.886 & 23.57 & 0.786 & 24.08 & 0.875 \\
LASIRR~(ICCV'21) & 23.34 & 0.812 & 24.36 & 0.898 & 23.72 & 0.903 & 25.73 & 0.902 & 23.45 & 0.808 & 24.18 & 0.893 \\
YTMT~(NeurIPS'21) & 23.26 & 0.806 & 24.87 & 0.896 & 22.91 & 0.884 & 25.48 & 0.890 & 23.85 & 0.810 & 24.04 & 0.880 \\
RobustSIRR~(CVPR'23) & 23.61 & 0.835 & 24.90 & 0.917 & 19.91 & 0.868 & 23.67 & 0.884 & 20.97 & 0.764 & 22.54 & 0.884 \\
DSRNet~(ICCV'23) & 23.91 & 0.818 & 26.74 & 0.920 & 24.83 & \underline{0.911} & 26.11 & 0.906 & 25.22 & 0.832 & 25.72 & 0.907 \\
RRW~(CVPR'24) & 23.82 & 0.817 & 26.55 & \underline{0.927} & 24.03 & 0.903 & 26.51 & 0.913 & 25.96 & 0.843 & 25.40 & 0.908 \\
DSIT~(NeurIPS'24) & 25.22 & 0.836 & \underline{27.27} & \textbf{0.932} & 25.58 & \textbf{0.922} & 27.40 & \underline{0.918} & 26.77 & \underline{0.847} & 26.50 & \textbf{0.919} \\
RDNet~(CVPR'25) & \underline{25.58} & \underline{0.846} & 26.78 & 0.921 & 26.33 & \textbf{0.922} & \underline{27.70} & 0.915 & 26.21 & 0.842 & 26.63 & 0.915 \\
DAI~(AAAI'26) & 25.24 & 0.840 & 27.26 & 0.920 & \underline{27.30} & \textbf{0.922} & 27.44 & 0.912 & \underline{27.05} & 0.846 & \underline{27.30} & \textbf{0.919} \\
\midrule
\rowcolor{oursbg}
\textbf{PRISM~(Ours)}
& \textbf{27.14} & \textbf{0.853}
& \textbf{27.61} & 0.925
& \textbf{27.85} & 0.906
& \textbf{29.54} & \textbf{0.932}
& \textbf{27.35} & \textbf{0.853}
& \textbf{27.95} & \underline{0.918} \\
\bottomrule
\end{tabular}
}
\end{table*}

\section{Experiment}

\subsection{Experimental Setup}

\noindent\textbf{Implementation Details.}
Our model is implemented in PyTorch and trained on a single NVIDIA A6000 GPU with a batch size of 2 and a crop size of $512\times512$ for 50{,}000 iterations.
We use AdamW~\cite{loshchilov2017decoupled} with cosine annealing~\cite{loshchilov2016sgdr}, decaying from $5 \times 10^{-5}$ to $1 \times 10^{-6}$, in bfloat16 precision.
We adopt FLUX.2 Klein 4B~\cite{blackforest2024flux} as the backbone.
Since our task does not involve text conditioning, we replace the text cross-attention input with a single zero-valued embedding token, effectively operating the transformer unconditionally; the text encoder is discarded to save memory.
Loss weights are $\lambda_{\text{lat}} = 1$, $\lambda_{\text{pix}} = 1$, $\lambda_{\text{lpips}} = 2$, $\lambda_{\text{cycle}} = 1$, $\lambda_{\text{LCS}} = 0.1$.
During training, paired images are augmented with random resize, horizontal flip, rotation, and a random $512\times512$ crop, applied consistently to input and ground truth.

\noindent\textbf{Training Data.}
We follow the same training data setting and synthesis pipeline as RDNet~\cite{zhao2025reversible}: 7{,}643 synthetic pairs from PASCAL VOC~\cite{everingham2010pascal}, 90 real pairs from~\cite{zhang2018single}, and 200 real pairs from Nature~\cite{li2020single}, sampled with a 0.6\,:\,0.2\,:\,0.2 ratio per epoch.
Real pairs provide additional diversity in reflection type and scene complexity not captured by the synthetic distribution.

\noindent\textbf{Evaluation Datasets.}
We evaluate PRISM on six standard benchmark datasets following the evaluation protocol of RDNet~\cite{zhao2025reversible}.
\textbf{Real}~\cite{zhang2018single} consists of 20 challenging real-world reflection images captured in diverse conditions.
\textbf{SIR$^2$}~\cite{wan2017benchmarking} is a benchmark comprising three subsets: Object (200 images), Postcard (199 images), and Wild (55 images), covering a wide range of scenes, illumination conditions, and glass types.
\textbf{Nature}~\cite{dong2021location} contains 20 real-world samples captured under natural lighting conditions.
For in-the-wild evaluation, we additionally use \textbf{OpenRR 1K}~\cite{yang2025openrr}, a real-world reflection removal benchmark; we evaluate on its test set of 100 images in an inference-only setting without any fine-tuning.

\noindent\textbf{Evaluation Metrics.}
We report PSNR and SSIM~\cite{wang2004image} on all benchmarks, following prior works~\cite{zhao2025reversible,hu2023single}, computed on the predicted transmission against ground truth in $[0, 1]$.
For OpenRR 1K, we additionally report LPIPS~\cite{zhang2018unreasonable} and DISTS~\cite{ding2020imagedists}, which measure perceptual similarity and correlate more strongly with human visual judgment, as well as NIQE~\cite{mittal2012making} as a no-reference image quality metric.

\begin{figure}[t]
    \centering
    \includegraphics[width=\linewidth]{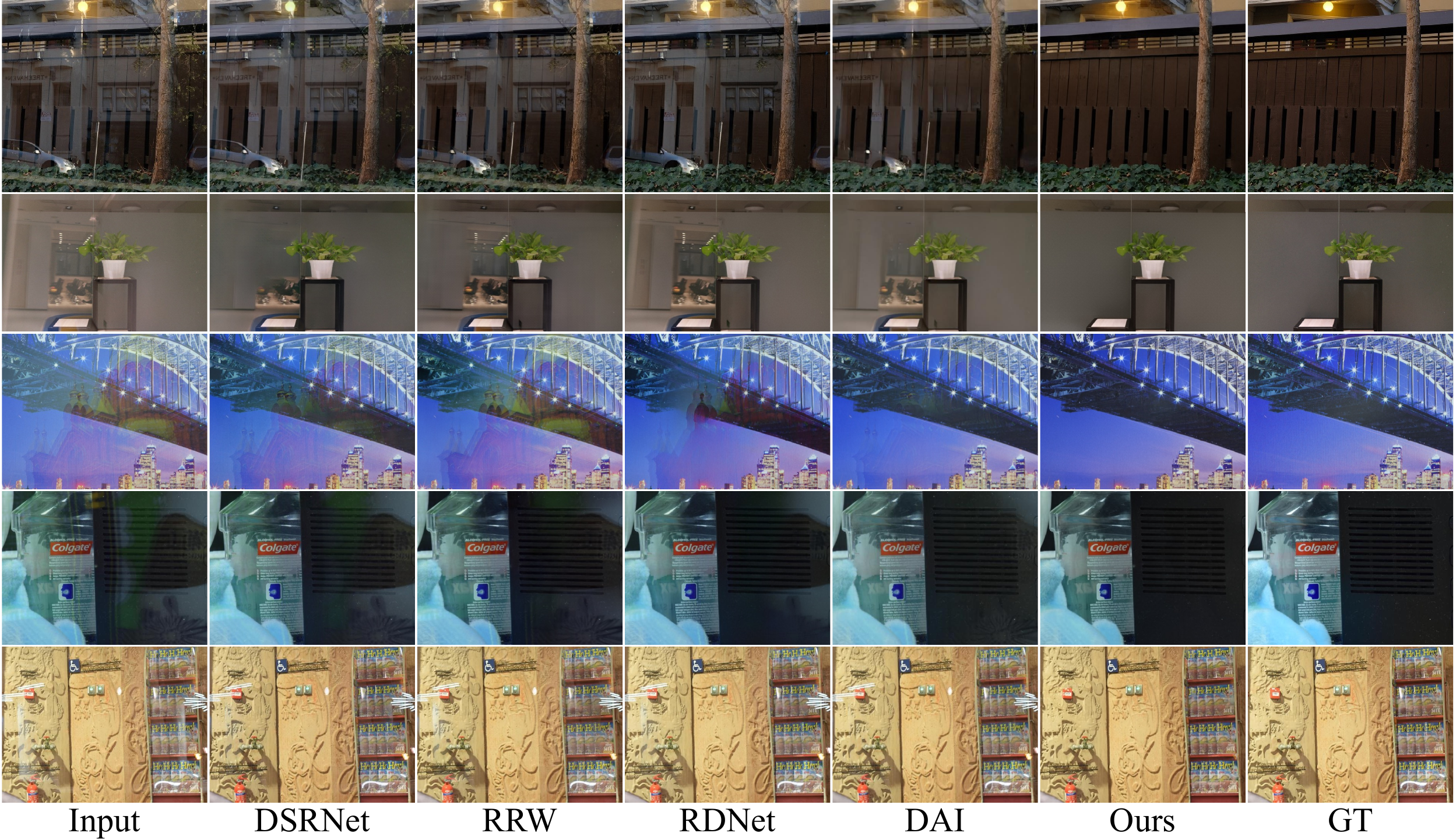}
    \caption{Visual comparison of different methods on multiple datasets. From top to bottom: Real, Nature, Postcard, Object, and Wild datasets. Zoom in for better visualization of detailed differences.}
    \label{fig:qualitative}
\end{figure}

\subsection{Comparison with State-of-the-Art Methods}

We compare PRISM against ten state-of-the-art SIRR approaches:
ERRNet~\cite{wei2019single}, IBCLN~\cite{li2020single}, LASIRR~\cite{dong2021location}, YTMT~\cite{hu2021trash}, RobustSIRR~\cite{kim2020single}, DSRNet~\cite{hu2023single}, RRW~\cite{zhu2024revisiting}, DSIT~\cite{hu2024single}, RDNet~\cite{zhao2025reversible}, and DAI~\cite{hu2025dereflection}.
In all tables, \textbf{bold} and \underline{underline} denote the best and second-best results, respectively.

\subsubsection{Quantitative Comparisons.}
As shown in Tab.~\ref{tab:benchmark_updated}, PRISM achieves state-of-the-art PSNR across all six benchmarks.
PRISM surpasses the previous best RDNet~\cite{zhao2025reversible} by \textbf{1.56 dB} on Real and \textbf{1.84 dB} on Wild, with consistent gains on all remaining benchmarks.
PRISM also outperforms the diffusion-based DAI~\cite{hu2025dereflection} on all reported datasets, demonstrating that structured latent decomposition provides a strong inductive bias for reflection removal that generalizes well across diverse real-world conditions.
On SSIM, PRISM achieves the best scores on Real, Wild, and Nature, while trailing DSIT~\cite{hu2024single} and DAI~\cite{hu2025dereflection} slightly on Object and Postcard.
This pattern is consistent with the known perception-distortion tradeoff~\cite{blau2018perception}: generative priors tend to recover sharper, higher-fidelity outputs that improve average pixel accuracy (PSNR) but may introduce local high-frequency details that deviate from the ground truth's exact spatial structure, which SSIM's window-based structural term penalizes.
This observation is further corroborated in Tab.~\ref{tab:in_the_wild_comparison}, where PRISM achieves substantially better perceptual scores (LPIPS $-14.0\%$, DISTS $-14.6\%$ relative to RDNet) despite slightly lower SSIM — metrics known to correlate more strongly with human visual judgment~\cite{zhang2018unreasonable,ding2020imagedists}.

\begin{figure}[t]
    \centering
    \includegraphics[width=\linewidth]{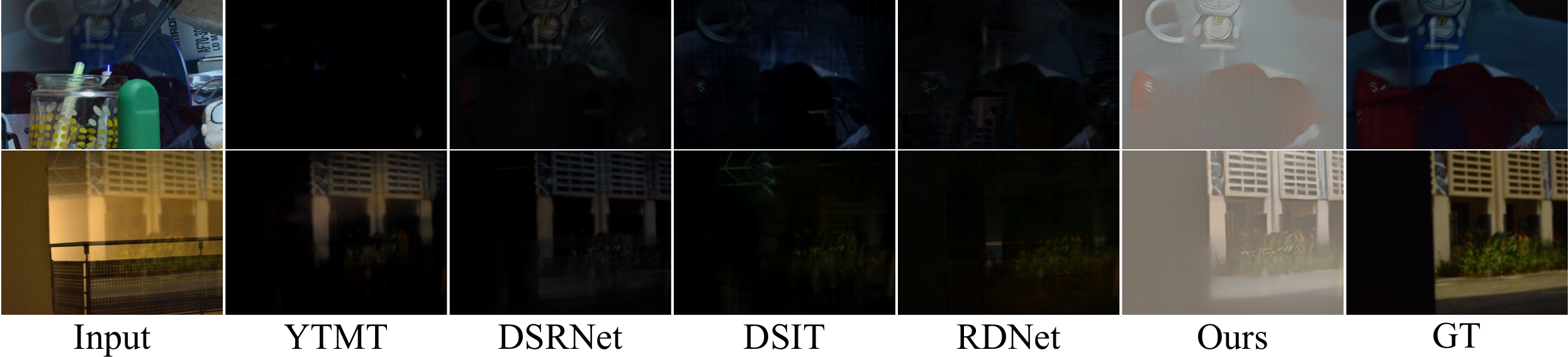}
    \caption{Qualitative comparison of estimated reflection layers. The top row shows a solid-object scene and the bottom row shows a wild scene, both from the SIR$^2$ dataset~\cite{wan2017benchmarking}. Best viewed zoomed in.}
    \label{fig:reflection}
\end{figure}

\subsubsection{Qualitative Comparisons.}
Fig.~\ref{fig:qualitative} shows visual results against DSRNet~\cite{hu2023single}, RRW~\cite{zhu2024revisiting},
RDNet~\cite{zhao2025reversible}, and DAI~\cite{hu2025dereflection} on five benchmarks.
Across all datasets, competing methods tend to either leave residual reflection 
artifacts or introduce color distortion and structural degradation in the recovered 
transmission.
Notably, in the last row, residual reflection patterns remain visible even in the
ground-truth image, whereas PRISM successfully removes these artifacts, yielding
a cleaner and more consistent result, further demonstrating the advantage of our
latent-space decomposition over pixel-space approaches.
Additional visual results are provided in the supplementary material.

Fig.~\ref{fig:reflection} compares the estimated reflection layers of YTMT~\cite{hu2021trash}, DSRNet~\cite{hu2023single}, DSIT~\cite{hu2024single}, and RDNet~\cite{zhao2025reversible} against ours.
All competing methods produce near-zero reflection estimates, indicating that pixel-space approaches tend to absorb the reflection component into reconstruction residuals rather than explicitly separating it.
This is a fundamental limitation of pixel-space decomposition: without a structured additive prior, the reflection branch degenerates into a trivial zero output while all information is passed to the transmission branch.
PRISM, by contrast, recovers structurally coherent reflections that preserve the spatial layout of the ground truth.
Although the recovered reflections exhibit brightness differences due to the absence of pixel-level reflection supervision, the clear structural correspondence demonstrates that latent-space decomposition achieves more effective layer disentanglement than pixel-space alternatives, even without a dedicated pixel-space reflection loss.

\subsubsection{In-the-wild Comparison.}
We further evaluate the generalization capability of PRISM on the in-the-wild OpenRR 1K dataset~\cite{yang2025openrr}.
As shown in Tab.~\ref{tab:in_the_wild_comparison}, all models are evaluated on the OpenRR 1K test set without being trained on this dataset, \ie, in an inference-only setting.
PRISM achieves the best performance in PSNR, LPIPS, DISTS, and NIQE among all compared methods.
While SSIM is lower than RDNet, PRISM outperforms all baselines on perceptual metrics (LPIPS $-14.0\%$, DISTS $-14.6\%$ relative to RDNet), consistent with the perception-distortion tradeoff discussed above.

Tab.~\ref{tab:in_the_wild_comparison} also reports the number of parameters and inference time.
Despite having more parameters, PRISM achieves comparable inference time to RDNet (104\,ms vs.\ 102\,ms), as \texttt{bf16} precision substantially reduces memory bandwidth and computation overhead.

\begin{figure}[t]
    \centering
    \includegraphics[width=\linewidth]{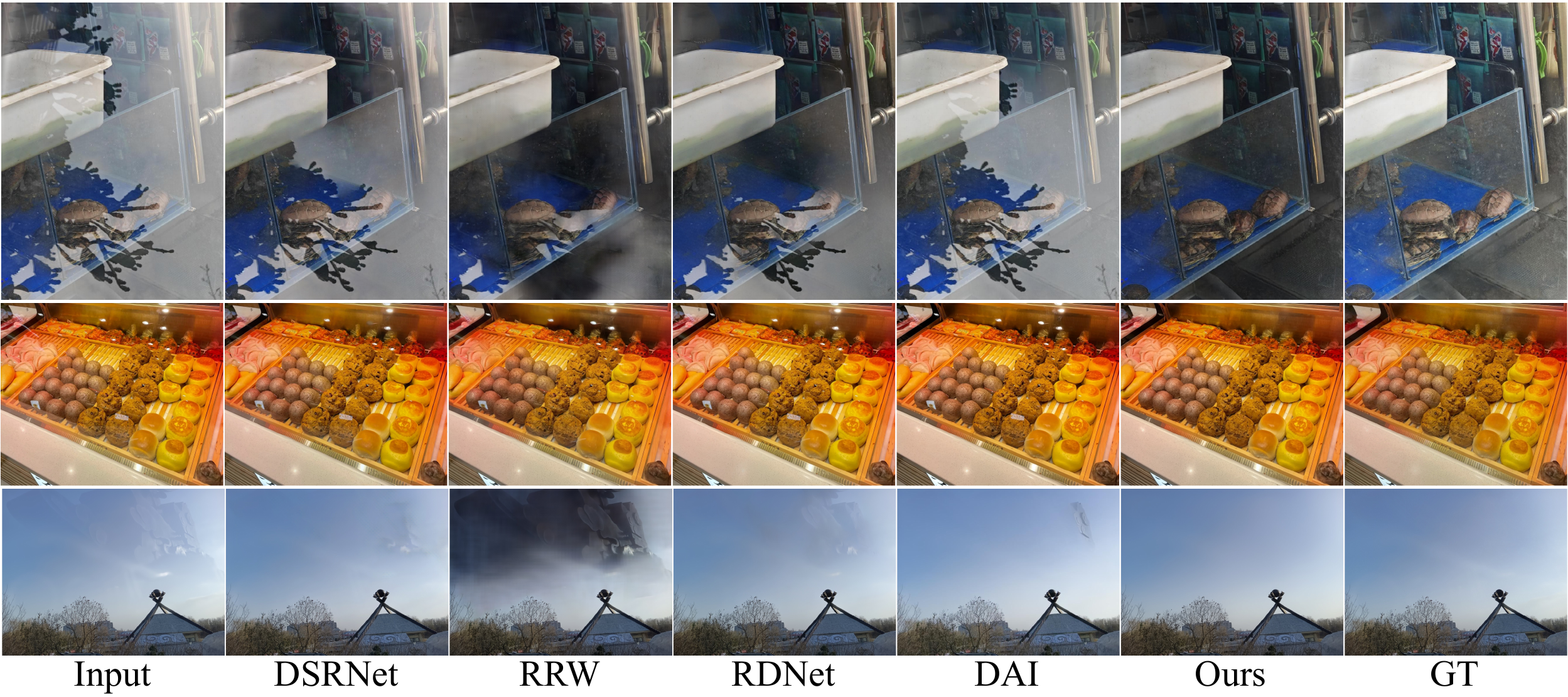}
    \caption{Visual comparison on the OpenRR 1K test set~\cite{yang2025openrr}. All methods are evaluated without training on this dataset.}
    \label{fig:in_the_wild}
\end{figure}

Fig.~\ref{fig:in_the_wild} provides visual comparisons.
PRISM effectively captures reflection patterns and removes them while preserving the underlying transmission structures.
Compared with competing methods, PRISM produces cleaner results with fewer residual artifacts, demonstrating stronger generalization in real-world reflection removal scenarios.

\begin{table}[t]
\centering
\caption{Quantitative comparison on the OpenRR 1K test set~\cite{yang2025openrr}. All methods are evaluated in an inference-only setting without training on this dataset. Inference is performed on $512\times512$ images, and inference time is measured on an NVIDIA RTX 4090 GPU.}
\label{tab:in_the_wild_comparison}
\setlength{\tabcolsep}{4pt}
\resizebox{0.9\linewidth}{!}{
\begin{tabular}{l|ccccc|cc}
\toprule
Method & PSNR$\uparrow$ & SSIM$\uparrow$ & LPIPS$\downarrow$ & DISTS$\downarrow$ & NIQE$\downarrow$ & Time (ms)$\downarrow$ & Params \\
\midrule
YTMT   & 25.16 & 0.927 & 0.096 & 0.059 & 3.6241 & \underline{75.68}  & \underline{58.47M} \\
DSRNet & 26.24 & \underline{0.935} & \underline{0.074} & \underline{0.048} & 3.6573 & 288.01 & 144.65M \\
DSIT   & 26.01 & 0.903 & 0.083 & 0.051 & \underline{3.5536} & 222.91 & 326.96M \\
RRW    & 25.55 & 0.929 & 0.086 & 0.055 & 3.6825 & \textbf{35.88} & \textbf{27.99M} \\
RDNet  & \underline{26.81} & \textbf{0.939} & \underline{0.074} & \underline{0.048} & 3.7713 & 102.26 & 314.03M \\
DAI    & 26.04 & 0.903 & 0.085 & 0.070 & 3.6197 & 138.70 & 1.30B \\
\midrule
\rowcolor{oursbg}
\textbf{PRISM} & \textbf{27.23} & 0.893 & \textbf{0.064} & \textbf{0.041} & \textbf{3.5512} & 104.23 & 3.97B \\
\bottomrule
\end{tabular}
}
\end{table}

\begin{table}[t]
\centering
\caption{Ablation study of $\mathcal{L}_{\text{cycle}}$ and $\mathcal{L}_{\text{LCS}}$. The base model (row 1) uses the velocity target $v^*\!=\!z_T\!-\!z_I$ supervised by $\mathcal{L}_{\text{latent}}$ and $\mathcal{L}_{\text{pixel}}$ only, without explicit disentanglement losses.}
\label{tab:ablation_cycle_nce}
\setlength{\tabcolsep}{4pt}
\resizebox{0.95\linewidth}{!}{
\begin{tabular}{cc|cc|cc|cc|cc}
\toprule
\multirow{2}{*}{$\mathcal{L}_{\text{cycle}}$} & \multirow{2}{*}{$\mathcal{L}_{\text{LCS}}$}
& \multicolumn{2}{c|}{\textbf{Real(20)}}
& \multicolumn{2}{c|}{\textbf{Nature(20)}}
& \multicolumn{2}{c|}{\textbf{SIR$^2$(454)}}
& \multicolumn{2}{c}{\textbf{Avg}} \\
\cmidrule(lr){3-4} \cmidrule(lr){5-6} \cmidrule(lr){7-8} \cmidrule(lr){9-10}
 &  & PSNR$\uparrow$ & SSIM$\uparrow$ & PSNR$\uparrow$ & SSIM$\uparrow$ & PSNR$\uparrow$ & SSIM$\uparrow$ & PSNR$\uparrow$ & SSIM$\uparrow$ \\
\midrule
$\times$     & $\times$     & 26.67 & \underline{0.847} & 27.24 & 0.850 & 27.68 & 0.916 & 27.20 & 0.871 \\
$\checkmark$ & $\times$     & \underline{26.90} & \textbf{0.853} & \textbf{27.35} & 0.851 & 27.86 & \underline{0.917} & \underline{27.37} & \underline{0.874} \\
$\times$     & $\checkmark$ & 26.74 & 0.846 & \underline{27.34} & \underline{0.852} & \underline{27.90} & \underline{0.917} & 27.33 & 0.872 \\
\rowcolor{oursbg}
$\checkmark$ & $\checkmark$ & \textbf{27.14} & \textbf{0.853} & \textbf{27.35} & \textbf{0.853} & \textbf{27.95} & \textbf{0.918} & \textbf{27.48} & \textbf{0.875} \\
\bottomrule
\end{tabular}
}
\end{table}

\begin{figure}[t]
    \centering
    \includegraphics[width=\linewidth]{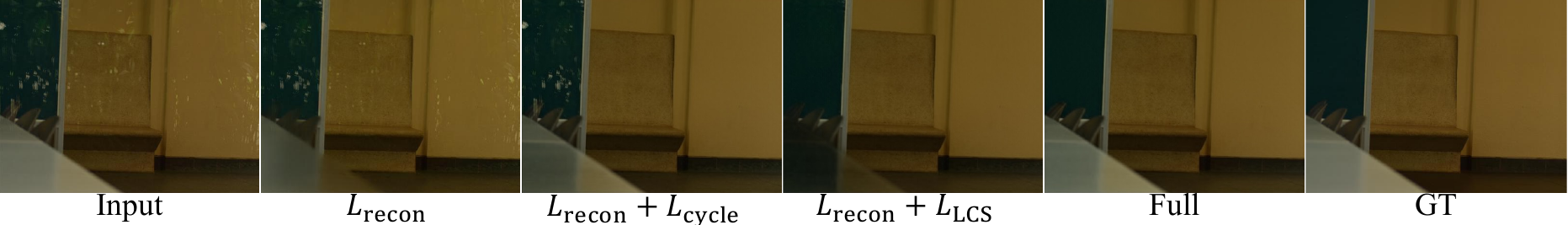}
    \caption{Progressive ablation of each loss component. $\mathcal{L}_\text{recon}{:=}\mathcal{L}_\text{latent}{+}\mathcal{L}_\text{pixel}$. Each column adds one loss on top of the previous setting.}
    \label{fig:ablation_qual}
\end{figure}

\begin{table}[t]
\centering
\caption{Ablation study on the cycle supervision target. $\tilde{z}_T$ and $\tilde{z}_R$ indicate whether the transmission and reflection components of the cycle branch are supervised, respectively.}
\label{tab:ablation_rt}
\setlength{\tabcolsep}{4pt}
\resizebox{0.95\linewidth}{!}{
\begin{tabular}{cc|cc|cc|cc|cc}
\toprule
\multirow{2}{*}{$\tilde{z}_T$} & \multirow{2}{*}{$\tilde{z}_R$}
& \multicolumn{2}{c|}{\textbf{Real(20)}}
& \multicolumn{2}{c|}{\textbf{Nature(20)}}
& \multicolumn{2}{c|}{\textbf{SIR$^2$(454)}}
& \multicolumn{2}{c}{\textbf{Avg}} \\
\cmidrule(lr){3-4} \cmidrule(lr){5-6} \cmidrule(lr){7-8} \cmidrule(lr){9-10}
 & & PSNR$\uparrow$ & SSIM$\uparrow$ & PSNR$\uparrow$ & SSIM$\uparrow$ & PSNR$\uparrow$ & SSIM$\uparrow$ & PSNR$\uparrow$ & SSIM$\uparrow$ \\
\midrule
$\times$ & $\checkmark$ & \underline{26.95} & \underline{0.850} & 27.26 & 0.851 & \underline{27.87} & \underline{0.917} & \underline{27.36} & \underline{0.873} \\
$\checkmark$ & $\times$ & 26.86 & 0.849 & \underline{27.32} & \underline{0.852} & 27.83 & \underline{0.917} & 27.34 & 0.872 \\
\rowcolor{oursbg}
$\checkmark$ & $\checkmark$ & \textbf{27.14} & \textbf{0.853} & \textbf{27.35} & \textbf{0.853} & \textbf{27.95} & \textbf{0.918} & \textbf{27.48} & \textbf{0.875} \\
\bottomrule
\end{tabular}
}
\end{table}

\begin{table}[t]
\centering
\setlength{\tabcolsep}{6pt}
\caption{Ablation study on spatial representation strategy for the Layer Contrastive Separation loss. Patch ($p$) denotes non-overlapping patch-average-pool with patch size $p$; Global Avg. Pool collapses the entire spatial map to a single vector.}
\label{tab:ablation_patch}
\resizebox{0.99\linewidth}{!}{
\begin{tabular}{c|cc|cc|cc|cc}
\toprule
Representation
& \multicolumn{2}{c|}{\textbf{Real(20)}}
& \multicolumn{2}{c|}{\textbf{Nature(20)}}
& \multicolumn{2}{c|}{\textbf{SIR$^2$(454)}}
& \multicolumn{2}{c}{\textbf{Avg}} \\
\cmidrule(lr){2-3} \cmidrule(lr){4-5} \cmidrule(lr){6-7} \cmidrule(lr){8-9}
& PSNR & SSIM
& PSNR & SSIM
& PSNR & SSIM
& PSNR & SSIM \\
\midrule
Patch ($p=2$)
& \underline{27.03} & \underline{0.851}
& \textbf{27.35} & \textbf{0.853}
& \underline{27.94} & \underline{0.917}
& \underline{27.44} & \underline{0.874} \\

\rowcolor{oursbg}
Patch ($p=4$, PRISM)
& \textbf{27.14} & \textbf{0.853}
& \textbf{27.35} & \textbf{0.853}
& \textbf{27.95} & \textbf{0.918}
& \textbf{27.48} & \textbf{0.875} \\

Patch ($p=8$)
& 26.83 & 0.849
& \underline{27.22} & \underline{0.852}
& 27.93 & \underline{0.917}
& 27.33 & 0.873 \\

Global Avg. Pool
& 26.73 & 0.845
& 27.21 & \underline{0.852}
& \underline{27.94} & \textbf{0.918}
& 27.29 & 0.872 \\
\bottomrule
\end{tabular}
}
\end{table}

\subsection{Ablation Study}

We conduct ablation studies on Real(20), Nature(20), and SIR$^2$(454) to analyze each proposed component.

\subsubsection{Effect of $\mathcal{L}_{\text{cycle}}$ and $\mathcal{L}_{\text{LCS}}$.}
Tab.~\ref{tab:ablation_cycle_nce} isolates the contribution of each disentanglement objective.
The base model (row~1) uses only $\mathcal{L}_{\text{latent}}$ and $\mathcal{L}_{\text{pixel}}$, confirming that the additive latent formulation provides a useful inductive bias on its own; however, without explicit disentanglement losses, the model may still entangle scene-specific information across the two components.
Adding $\mathcal{L}_{\text{cycle}}$ yields consistent improvements across all three benchmarks (+0.23\,dB on Real, +0.18\,dB on SIR$^2$), confirming that enforcing reconstruction consistency under cyclically swapped reflections strengthens disentanglement.
Combining both losses further improves performance, with the full model achieving 27.14\,/\,27.35\,/\,27.95\,dB on Real, Nature, and SIR$^2$ respectively, demonstrating that $\mathcal{L}_{\text{cycle}}$ and $\mathcal{L}_{\text{LCS}}$ are complementary: the cycle loss enforces structural consistency in latent space, while $\mathcal{L}_{\text{LCS}}$ provides a semantic disentanglement signal without relying on exact additive reconstruction.
Fig.~\ref{fig:ablation_qual} visualizes the progressive effect: $\mathcal{L}_\text{recon}$ alone leaves residual reflections; adding $\mathcal{L}_\text{cycle}$ stabilizes the transmission but reflections persist; $\mathcal{L}_\text{LCS}$ suppresses them further yet introduces minor artifacts; only the full combination removes reflections while preserving fidelity.

\subsubsection{Effect of Cycle Supervision Target.}
Tab.~\ref{tab:ablation_rt} investigates which components are supervised within the cycle branch.
Supervising only the reflection cycle ($\tilde{z}_R$) or only the transmission cycle ($\tilde{z}_T$) in isolation both degrade performance compared to supervising both jointly, underscoring the importance of bidirectional consistency.
Jointly supervising both components achieves the best results across all benchmarks, as each direction provides a complementary constraint: the transmission cycle enforces that the scene content is preserved under arbitrary reflection corruption, while the reflection cycle ensures that the separated reflection latent is consistent with the physical composition.

\subsubsection{Effect of Patch-Based Representation.}
Tab.~\ref{tab:ablation_patch} compares spatial representation strategies for $\mathcal{L}_{\text{LCS}}$.
Global average pooling performs worst (26.73\,dB on Real), as collapsing the entire latent map into a single vector discards the spatially heterogeneous entanglement structure.
Patch-based representations consistently outperform global pooling across all tested sizes, with $p\!=\!4$ achieving the best balance between spatial granularity and representation stability (27.14\,/\,27.95\,dB on Real\,/\,SIR$^2$).
Too-small patches ($p\!=\!2$) introduce noise from overly local comparisons, while too-large patches ($p\!=\!8$) lose the fine-grained spatial cues needed to distinguish locally entangled layers.

\section{Conclusion}

In this paper, we presented PRISM, a latent flow matching framework for single-image reflection removal that reinterprets the task as a latent linear separation problem.
By exploiting the low mutual coherence of pretrained VAE latent representations, PRISM adopts an approximate additive formulation and recovers the transmission via a velocity field in a single forward pass, without dual-branch architectures or a dedicated pixel-space reflection loss.
To enforce robust disentanglement, PRISM introduces a Latent Composition Consistency strategy — uniquely enabled by the broad decorrelation of latent representations — and a Layer Contrastive Separation loss that handles the non-exact nature of the additive assumption without propagating approximation errors.
Extensive experiments on six real-world benchmarks demonstrate that PRISM consistently outperforms state-of-the-art approaches, achieving significant PSNR gains while maintaining strong generalization to challenging real-world imagery.

\subsubsection{Limitations.}
\begin{figure}[t]
    \centering
    \includegraphics[width=\linewidth]{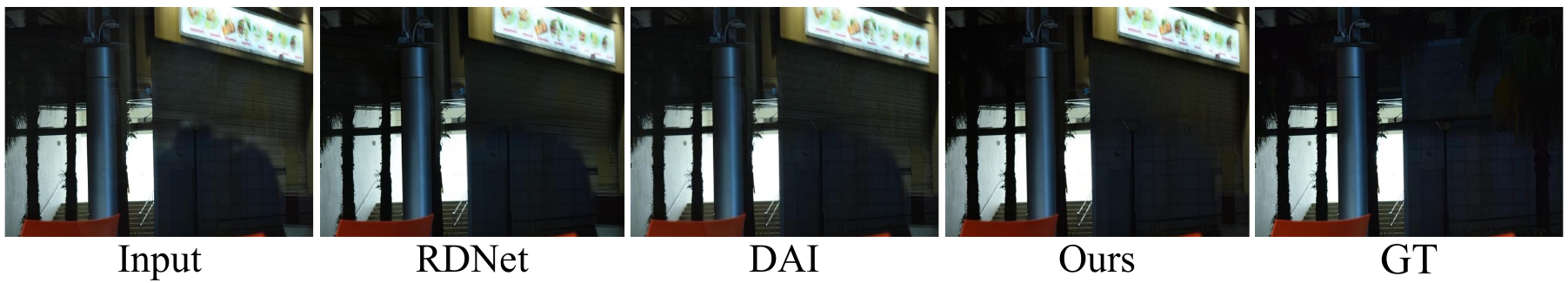}
    \caption{Failure case: dense and saturated reflection on SIR$^2$~\cite{wan2017benchmarking}. From left to right: input $I$, RDNet~\cite{zhao2025reversible}, DAI~\cite{hu2025dereflection}, Ours, and ground truth $T$.}
    \label{fig:failure}
\end{figure}

The FLUX backbone (3.97B parameters) incurs significant memory overhead, limiting deployment on resource-constrained hardware.
Additionally, while the additive latent formulation is empirically effective, a principled theoretical understanding of when and why the VAE encoder approximately preserves additive structure remains an open question.
Fig.~\ref{fig:failure} illustrates a representative failure mode: when the reflection is optically dense and saturated (\eg, $\|R\|_1 / \|I\|_1 > 0.5$), $z_I$ shifts toward the reflection subspace and the low-coherence assumption weakens, causing all methods—including PRISM—to leave residual reflections.
We provide further analysis in the supplementary material.

\section*{Acknowledgments}
This research was supported by Culture, Sports and Tourism R\&D Program through the Korea Creative Content Agency grant funded by the Ministry of Culture, Sports and Tourism in 2026 (Project Name: Development of AI-Based Animation Production Technology to Ensure Character and Scene Consistency and Continuity for Enhanced Efficiency and Quality, Project Number: RS-2026-25525207, Contribution Rate: 30\%). 
This work was also supported by Institute of Information \& communications Technology Planning \& Evaluation (IITP) grant funded by the Korea government (MSIT) (No.2022-0-00156, Fundamental research on continual meta-learning for quality enhancement of casual videos and their 3D metaverse transformation), 
IITP grant funded by the Korea government (MSIT) (No.RS-2020-II201373, Artificial Intelligence Graduate School Program (Hanyang University)). 

%
%
\bibliographystyle{splncs04}
\bibliography{main}


\clearpage

\appendix

\renewcommand{\thesection}{S\arabic{section}}
\renewcommand{\thefigure}{S\arabic{figure}}
\renewcommand{\thetable}{S\arabic{table}}

\setcounter{section}{0}
\setcounter{figure}{0}
\setcounter{table}{0}

\renewcommand{\theHsection}{supp.\thesection}
\renewcommand{\theHfigure}{supp.\thefigure}
\renewcommand{\theHtable}{supp.\thetable}

\title{PRISM: Latent Composition Consistency for \\ Single-Image Reflection Removal}


\begin{center}
  {\Large\bfseries PRISM: Latent Composition Consistency for Single-Image Reflection Removal}\\[1.0em]
  {\large\bfseries Supplementary Material}
\end{center}


\section*{Overview}

This supplementary material provides additional experiments and analyses to complement the main paper.

\begin{itemize}
    \item \textbf{Sec.~\ref{sec:supp:additivity}}: Empirical analysis of the latent additivity approximation, substantiating the key theoretical claim underlying PRISM.
    \item \textbf{Sec.~\ref{sec:supp:backbone}}: Lightweight backbone ablation (SD-2.1) to contextualize the role of model scale.
    \item \textbf{Sec.~\ref{sec:supp:reflection}}: Post-hoc reflection calibration analysis to validate the structural quality of recovered reflections.
    \item \textbf{Sec.~\ref{sec:supp:lossweight}}: Sensitivity analysis for loss weights $\lambda_{\text{cycle}}$ and $\lambda_{\text{LCS}}$.
    \item \textbf{Sec.~\ref{sec:supp:sg}}: Stop-gradient ablation for the cycle consistency loss.
    \item \textbf{Sec.~\ref{sec:supp:timestep}}: Timestep sampling strategy comparison ($t\!=\!0$ vs.\ $t \sim \mathcal{U}[0,1]$).
    \item \textbf{Sec.~\ref{sec:supp:discussion}}: Discussion of design choices: trivial solution avoidance and batch size sensitivity.
    \item \textbf{Sec.~\ref{sec:supp:qualitative}}: Additional qualitative comparisons across all benchmarks.
    \item \textbf{Sec.~\ref{sec:supp:generative}}: Visual comparison between generative-prior-based approaches (DAI and PRISM) on real-world images.
    \item \textbf{Sec.~\ref{sec:supp:pixelablation}}: Pixel-only and pixel-cycle ablations to isolate the contribution of latent-space supervision.
    \item \textbf{Sec.~\ref{sec:supp:failure}}: Additional failure case analysis (structural glass artifacts).
\end{itemize}

\section{Empirical Latent Additivity Analysis}
\label{sec:supp:additivity}

\subsubsection{Motivation.}
A natural concern about PRISM is the following apparent contradiction:
the main paper reports a scale-normalized additivity error of $\bar{\varepsilon} = 0.9235$ in latent space,
substantially higher than $\bar{\varepsilon} = 0.2673$ in pixel space.
If the additive constraint $z_I \approx z_T + z_R$ holds \emph{worse} in latent space,
why perform decomposition there?
This section resolves the apparent contradiction by showing that
\emph{additivity error and directional coherence are orthogonal axes}:
a high $\varepsilon$ does not imply high coherence,
and it is low coherence — not additive exactness — that determines whether a linear decomposition is feasible.

\begin{figure}[t]
    \centering
    \includegraphics[width=0.82\linewidth]{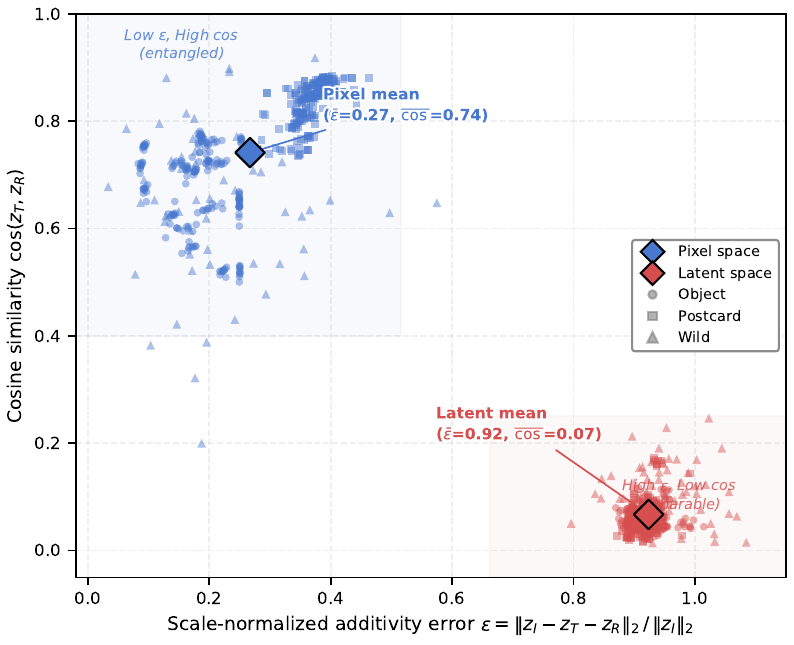}
    \caption{Additivity error $\varepsilon$ vs.\ cosine similarity $\cos(z_T, z_R)$ for all 454 SIR$^2$~\cite{wan2017benchmarking} pairs,
    computed in pixel space (\textcolor[rgb]{0.28,0.47,0.81}{$\circ$}, blue)
    and FLUX~\cite{blackforest2024flux} VAE latent space (\textcolor[rgb]{0.84,0.37,0.34}{$\times$}, red).
    Diamond markers indicate per-space means.}
    \label{fig:supp:additivity}
\end{figure}

\subsubsection{Analysis.}
We compute two complementary metrics for all 454 SIR$^2$~\cite{wan2017benchmarking} pairs in both pixel and latent space:
\begin{itemize}
    \item \textbf{Additivity error} $\varepsilon = \|z_I - z_T - z_R\|_2 / \|z_I\|_2$:
          measures residual energy after the additive decomposition, normalized by input magnitude.
    \item \textbf{Cosine similarity} $\cos(z_T, z_R) = z_T^\top z_R / (\|z_T\|_2 \|z_R\|_2)$:
          a scale-invariant measure of directional overlap between the two components.
\end{itemize}

Fig.~\ref{fig:supp:additivity} plots these two metrics against each other for every pair in both spaces.
The two point clouds occupy \emph{opposite quadrants}:
pixel-space representations cluster at low $\varepsilon$ ($\bar{\varepsilon} = 0.27$) but high cosine similarity ($\overline{\cos} = 0.74$),
while latent-space representations cluster at high $\varepsilon$ ($\bar{\varepsilon} = 0.92$) but near-zero cosine similarity ($\overline{\cos} = 0.07$).
Crucially, this pattern is consistent across all scene categories:
the latent cosine similarity remains near zero for Object ($0.058$), Postcard ($0.064$), and Wild ($0.112$),
while the corresponding pixel-space values are substantially higher ($0.677$, $0.835$, $0.639$).

\subsubsection{Interpretation.}
This result reveals a fundamental trade-off between the two spaces.
In pixel space, the additive model holds well ($\varepsilon = 0.27$).
However, $T$ and $R$ point in similar directions ($\cos = 0.74$),
meaning that a linear decomposition cannot reliably distinguish them —
the two components are entangled in the direction sense that matters for separation.

In latent space, the VAE encoder's nonlinearity breaks the additive structure ($\varepsilon \approx 0.92$),
so $z_I \approx z_T + z_R$ is only a working approximation.
Yet the encoder simultaneously reduces the mutual coherence between $z_T$ and $z_R$ ($\cos \leq 0.112$ across all categories),
making the two components \emph{directionally} distinguishable.
We note that random unit vectors in high-dimensional spaces also exhibit near-zero expected cosine similarity ($\mathbb{E}[\cos] \to 0$ as $d \to \infty$); however, the pixel-space vectors share the same dimensionality yet yield $\overline{\cos} = 0.74$, demonstrating that the low latent-space cosine is a property of the \emph{encoder mapping}, not merely a dimensional artifact.
It is this low coherence — not additive exactness — that allows PRISM to learn a reliable
velocity field $v_\theta$ that separates the two layers.
This is empirically supported by the ablation in Tab.~3 of the main paper: even the base model (without LCC/LCS) achieves 26.67\,dB on Real, indicating that the latent-space formulation alone provides a strong inductive bias for decomposition.
Furthermore, the tight clustering of the latent points ($\sigma_\varepsilon = 0.032$, $\sigma_{\cos} = 0.035$)
confirms that this property is structurally stable, not an artifact of particular images or scene categories.

\subsubsection{Cross-image control experiment.}
A natural question is whether the low cosine similarity in latent space is specific to the $(T, R)$ pairing or a general property of any two encoded images.
To test this, we compute $\cos(T, T')$ and $\cos(T, R')$, where $T'$ and $R'$ denote the transmission and reflection from a \emph{different} sample in the dataset.

\begin{table}[h]
\centering
\caption{Cosine similarity of latent components on SIR$^2$~\cite{wan2017benchmarking}. $T'$/$R'$ denote transmission and reflection from a different sample (cross-image control).}
\label{tab:supp:cos_control}
\setlength{\tabcolsep}{4pt}
\resizebox{\linewidth}{!}{
\begin{tabular}{l|ccc|c}
\toprule
Space & $\cos(T,R)$ & $\cos(T,T')$ & $\cos(T,R')$ & $1/\sqrt{d}$ \\
\midrule
Pixel       & 0.741 & 0.745 & 0.736 & 0.001 \\
\midrule
SD2.1       & 0.248 & 0.157 & 0.223 & 0.008 \\
SD3         & 0.596 & 0.544 & 0.565 & 0.004 \\
\rowcolor{oursbg}
FLUX        & \textbf{0.067} & \textbf{0.072} & \textbf{0.056} & 0.003 \\
\bottomrule
\end{tabular}}
\end{table}

Table~\ref{tab:supp:cos_control} reveals two key findings.
First, in FLUX latent space, $\cos(T, T') = 0.072$ is close to $\cos(T, R) = 0.067$, confirming that the decorrelation is \emph{broad} rather than specific to the transmission-reflection pairing.
The FLUX VAE does not inherently separate transmission from reflection; rather, it maps any pair of natural images into a low-coherence representation space.
Second, this drop is encoder-induced, not a dimensionality artifact: pixel space is \emph{higher}-dimensional ($d = 786$K) yet retains $\cos \approx 0.74$, while the expected cosine for random vectors ($1/\sqrt{d}$) is far below the observed pixel-space values.
The practical consequence is that in pixel space $T$ and $R$ overlap heavily ($\cos = 0.74$), making regression from $I$ to $T$ ill-conditioned, whereas in FLUX latent space the overlap vanishes and the regression becomes substantially easier.
This is further corroborated by the SIRR PSNR trend across VAE encoders: switching from SD-2.1 to FLUX yields $+3.17$\,dB (Tab.~\ref{tab:supp:backbone}), tracking the magnitude of the cosine similarity drop.

\section{Lightweight Backbone Ablation}
\label{sec:supp:backbone}

\subsubsection{Motivation.}
PRISM adopts a 3.97B-parameter FLUX~\cite{blackforest2024flux} backbone.
To verify that the performance gains are attributable to the proposed LCC and LCS objectives rather than to raw model capacity,
we fine-tune a smaller backbone with the same training strategy and compare against the full PRISM model.

\subsubsection{Setup.}
We adopt a Stable Diffusion 2.1~\cite{rombach2022high} U-Net backbone ($\sim$860M parameters) and apply identical LCC and LCS objectives,
training data, and hyperparameters as the main model.
Although SD-2.1 is originally a denoising diffusion model rather than a flow matching model, we repurpose its U-Net as a direct regression network: given the input latent $z_I$, the network predicts the velocity $v_\theta(z_I, 0)$ with the same $t\!=\!0$ training objective used by PRISM, bypassing the diffusion noise schedule entirely.
We compare three configurations: (i) SD-2.1 without LCC/LCS, (ii) SD-2.1 with LCC+LCS, and (iii) full PRISM (FLUX + LCC+LCS).

\begin{table}[h]
\centering
\caption{Lightweight backbone ablation on Real~\cite{zhang2018single}, Nature~\cite{li2020single}, and SIR$^2$~\cite{wan2017benchmarking}(454). All models are trained with identical data and loss weights. Avg denotes the sample-weighted average across all 494 test images: $\text{Avg} = (20\!\times\!\text{Real} + 20\!\times\!\text{Nature} + 454\!\times\!\text{SIR}^2)/494$.}
\label{tab:supp:backbone}
\setlength{\tabcolsep}{6pt}
\resizebox{\linewidth}{!}{
\begin{tabular}{l|c|cc|cc|cc|cc}
\toprule
\multirow{2}{*}{Model} & \multirow{2}{*}{Params}
& \multicolumn{2}{c|}{\textbf{Real(20)}}
& \multicolumn{2}{c|}{\textbf{Nature(20)}}
& \multicolumn{2}{c|}{\textbf{SIR$^2$(454)}}
& \multicolumn{2}{c}{\textbf{Avg}} \\
& & PSNR$\uparrow$ & SSIM$\uparrow$ & PSNR$\uparrow$ & SSIM$\uparrow$ & PSNR$\uparrow$ & SSIM$\uparrow$ & PSNR$\uparrow$ & SSIM$\uparrow$ \\
\midrule
SD-2.1 (baseline, no LCC/LCS)   & 860M  & 23.10 & 0.692 & 26.56 & 0.793 & 24.40 & 0.826 & 24.34 & 0.818 \\
SD-2.1 + LCC + LCS              & 860M  & 23.36 & 0.692 & 26.40 & 0.790 & 24.84 & 0.828 & 24.73 & 0.820 \\
\midrule
\rowcolor{oursbg}
PRISM (FLUX + LCC + LCS)        & 3.97B & \textbf{27.14} & \textbf{0.853} & \textbf{27.35} & \textbf{0.853} & \textbf{27.95} & \textbf{0.918} & \textbf{27.90} & \textbf{0.913} \\
\bottomrule
\end{tabular}
}
\end{table}

\subsubsection{Analysis.}
Table~\ref{tab:supp:backbone} shows that adding LCC+LCS to SD-2.1 improves SIR$^2$ PSNR by $+$0.44\,dB and Real PSNR by $+$0.26\,dB over the SD-2.1 baseline.
Nature shows a marginal drop ($-$0.16\,dB), likely because the SD-2.1 latent space exhibits weaker decorrelation than FLUX (see Sec.~\ref{sec:supp:additivity}), limiting the effectiveness of LCC/LCS on certain scene types.
Nevertheless, the overall trend confirms that the proposed objectives provide a consistent gain even on a much smaller backbone.
The absolute performance gap between SD-2.1 and PRISM ($\sim$3\,dB) highlights the additional benefit of the FLUX backbone's stronger generative prior and larger model capacity.
This finding is consistent with recent observations in other image restoration tasks, where the choice of backbone and its underlying degradation modeling significantly affect performance~\cite{shin2025hazeflow,im2023deep}.

\section{Post-hoc Reflection Calibration}
\label{sec:supp:reflection}

\subsubsection{Motivation.}
Unlike existing methods that directly supervise the reflection layer with pixel-space losses (\eg, $\|\hat{R} - R_\text{gt}\|_1$), PRISM recovers reflections purely through latent-space objectives: the flow matching target $\hat{z}_R = -v_\theta$ and the LCC/LCS losses.
As a result, the decoded reflections $\hat{R} = \mathcal{D}(\hat{z}_R)$ exhibit a systematic brightness bias---regions where $R_\text{gt} \approx 0$ do not reach zero in $\hat{R}$, while strongly reflective regions are predicted at roughly correct intensity.
We show that this deviation is a global affine shift rather than a structural failure:
after a simple per-image affine calibration, PRISM achieves reflection recovery quality competitive with methods that use direct pixel-space reflection supervision, confirming that the latent flow matching framework captures the correct spatial structure of reflections.

\subsubsection{Calibration procedure.}
Given a recovered reflection $\hat{R}$ and its ground truth $R_\text{gt}$, we apply a per-image affine transformation
$\hat{R}_\text{cal} = \alpha^* \hat{R} + \beta^*$,
where $\alpha^*$ and $\beta^*$ are obtained by ordinary least-squares regression of $\hat{R}$ onto $R_\text{gt}$:
\begin{equation}
    \alpha^* = \frac{\mathrm{Cov}(\hat{R},\, R_\text{gt})}{\mathrm{Var}(\hat{R})}, \qquad
    \beta^* = \mu_{R_\text{gt}} - \alpha^* \mu_{\hat{R}},
\end{equation}
where $\mu$ denotes the spatial mean.
The scale term $\alpha^*$ corrects the global brightness, while the offset term $\beta^*$ removes a systematic positive bias that prevents reflection-free regions from reaching zero.
This calibration does not modify the transmission prediction and is applied only for the reflection quality evaluation.

\begin{table}[h]
\centering
\caption{Reflection recovery quality before and after post-hoc affine calibration on SIR$^2$~\cite{wan2017benchmarking}(454). $\hat{R}$: uncalibrated output; $\hat{R}_\text{cal}$: after per-image affine calibration. All methods are evaluated under the same protocol.}
\label{tab:supp:reflection}
\setlength{\tabcolsep}{4pt}
\resizebox{\linewidth}{!}{
\begin{tabular}{l|cc|cc|cc|cc}
\toprule
\multirow{2}{*}{Method}
& \multicolumn{2}{c|}{\textbf{Object(200)}}
& \multicolumn{2}{c|}{\textbf{Postcard(199)}}
& \multicolumn{2}{c|}{\textbf{Wild(55)}}
& \multicolumn{2}{c}{\textbf{SIR$^2$(454)}} \\
& PSNR$\uparrow$ & SSIM$\uparrow$ & PSNR$\uparrow$ & SSIM$\uparrow$ & PSNR$\uparrow$ & SSIM$\uparrow$ & PSNR$\uparrow$ & SSIM$\uparrow$ \\
\midrule
YTMT~\cite{hu2021trash} $\hat{R}$                     & 18.77 & 0.062 & 11.83 & 0.144 & 20.18 & 0.196 & 15.90 & 0.114 \\
YTMT $\hat{R}_\text{cal}$          & 24.72 & 0.778 & 18.24 & 0.646 & 24.60 & 0.737 & 21.86 & 0.715 \\
\midrule
DSRNet~\cite{hu2023single} $\hat{R}$  & 20.62 & 0.485 & 12.20 & 0.280 & 22.00 & 0.542 & 17.10 & 0.402 \\
DSRNet $\hat{R}_\text{cal}$        & 25.18 & 0.786 & 18.36 & 0.638 & 25.34 & 0.752 & 22.21 & 0.717 \\
\midrule
DSIT~\cite{hu2024single} $\hat{R}$    & 21.73 & 0.489 & 13.10 & 0.311 & 22.02 & 0.493 & 17.98 & 0.411 \\
DSIT $\hat{R}_\text{cal}$          & \textbf{25.75} & \textbf{0.793} & \textbf{19.24} & \textbf{0.676} & 25.84 & \textbf{0.765} & \textbf{22.90} & \textbf{0.738} \\
\midrule
RDNet~\cite{zhao2025reversible} $\hat{R}$  & 21.30 & 0.467 & 13.06 & 0.273 & 22.07 & 0.518 & 17.78 & 0.388 \\
RDNet $\hat{R}_\text{cal}$         & 25.53 & 0.771 & 18.59 & 0.625 & 25.41 & 0.730 & 22.48 & 0.702 \\
\midrule
\rowcolor{oursbg}
PRISM $\hat{R}$                    & 8.27  & 0.229 & 10.56 & 0.404 & 8.13  & 0.188 & 9.26  & 0.301 \\
\rowcolor{oursbg}
PRISM $\hat{R}_\text{cal}$         & 25.46 & 0.705 & 19.26 & 0.500 & \textbf{26.25} & 0.671 & 22.83 & 0.611 \\
\bottomrule
\end{tabular}
}
\end{table}

\subsubsection{Quantitative analysis.}
Table~\ref{tab:supp:reflection} reports the reflection recovery quality before and after affine calibration.
PRISM exhibits the lowest uncalibrated PSNR (9.26\,dB) because it receives no pixel-space reflection loss, yet achieves the largest calibration gain of $+$13.57\,dB, reaching 22.83\,dB after a simple two-parameter correction.
This places PRISM within 0.07\,dB of DSIT~\cite{hu2024single} (22.90\,dB), the strongest baseline, which is trained with an explicit $\|\hat{R} - R_\text{gt}\|_2^2$ loss.
Notably, all methods benefit substantially from affine calibration (4--6\,dB gain), confirming that the protocol provides a fairer reflection evaluation by decoupling structural fidelity from global brightness alignment.
We emphasize that PRISM's disproportionately large calibration gain ($+$13.57\,dB vs.\ 4--6\,dB for baselines) reflects the absence of pixel-space reflection supervision, not a structural advantage of the calibration protocol; the purpose of this analysis is to demonstrate that the latent flow matching framework captures the correct \emph{spatial structure} of reflections despite lacking direct brightness supervision, rather than to claim competitive uncalibrated reflection quality.

\subsubsection{Qualitative comparison.}
\begin{figure}[h]
    \centering
    \includegraphics[width=\linewidth]{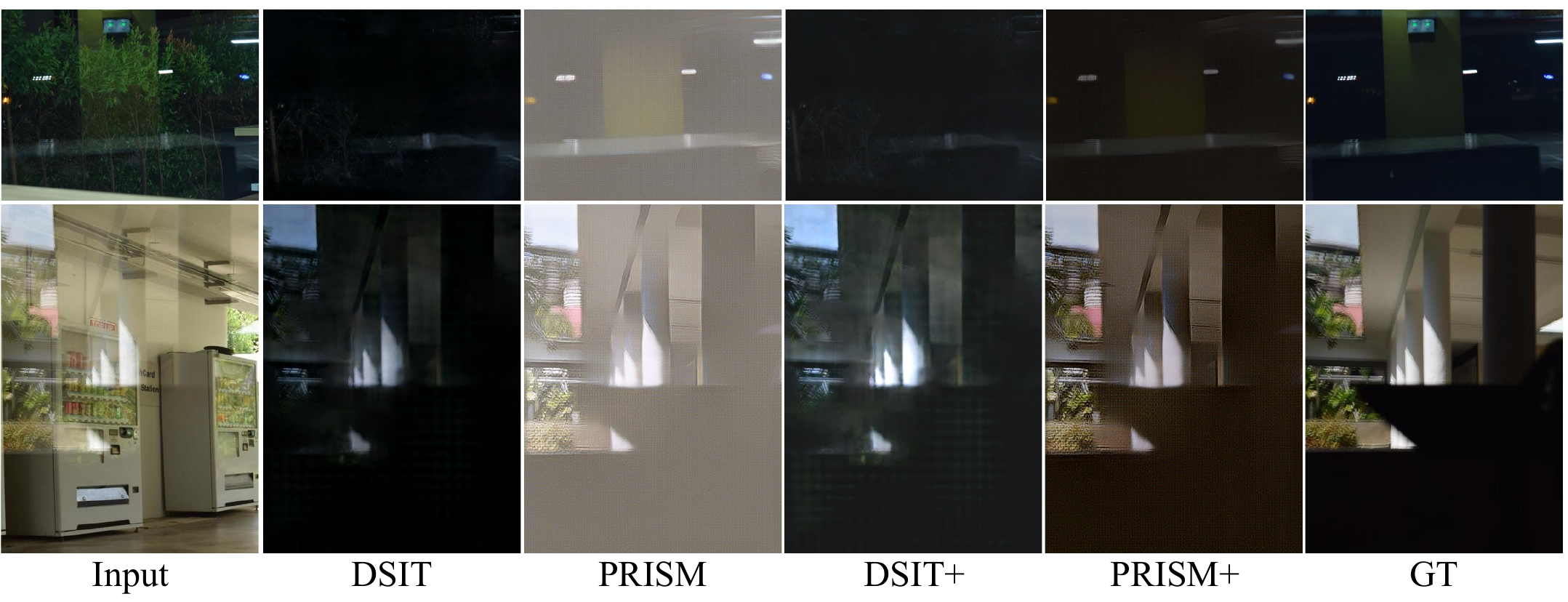}
    \caption{Post-hoc affine calibration of estimated reflection layers on SIR$^2$~\cite{wan2017benchmarking}.
    From left to right: input mixture $I$, uncalibrated DSIT~\cite{hu2024single} $\hat{R}$, uncalibrated PRISM $\hat{R}$, calibrated DSIT+ $\hat{R}_\text{cal}$, calibrated PRISM+ $\hat{R}_\text{cal}$, and ground-truth reflection $R_\text{gt}$.}
    \label{fig:supp:reflection}
\end{figure}

Fig.~\ref{fig:supp:reflection} compares PRISM and DSIT~\cite{hu2024single} (the strongest baseline after calibration) before and after affine calibration.
The uncalibrated PRISM reflections appear uniformly bright due to the absence of direct pixel-space reflection supervision, yet the underlying spatial structure---edges, textures, and object boundaries---is clearly preserved.
After applying the two-parameter affine correction, PRISM+ recovers reflections that are visually comparable to DSIT+, despite DSIT being trained with an explicit $\|\hat{R} - R_\text{gt}\|_2^2$ loss.
This confirms that the brightness deviation in PRISM is a simple global shift that can be removed post-hoc, rather than a fundamental limitation of the latent decomposition.

\section{Loss Weight Sensitivity}
\label{sec:supp:lossweight}

\subsubsection{Motivation.}
PRISM uses five loss weights: $\lambda_{\text{lat}} = 1$, $\lambda_{\text{pix}} = 1$, $\lambda_{\text{lpips}} = 2$, $\lambda_{\text{cycle}} = 1$, and $\lambda_{\text{LCS}} = 0.1$.
We verify that performance is robust to these choices by independently varying two key weights — $\lambda_{\text{cycle}}$ and $\lambda_{\text{LCS}}$ — while keeping all others fixed.
$\lambda_{\text{LCS}}$ is set to a smaller value by default because the InfoNCE~\cite{oord2018representation} loss operates on a different scale than the regression losses; we confirm below that this choice is not critically sensitive.

\begin{table}[h]
\centering
\caption{Sensitivity to cycle loss weight $\lambda_{\text{cycle}}$. All other weights are fixed at their default values.}
\label{tab:supp:lw_cycle}
\setlength{\tabcolsep}{5pt}
\resizebox{\linewidth}{!}{
\begin{tabular}{c|cc|cc|cc|cc}
\toprule
\multirow{2}{*}{$\lambda_{\text{cycle}}$}
& \multicolumn{2}{c|}{\textbf{Real(20)}}
& \multicolumn{2}{c|}{\textbf{Nature(20)}}
& \multicolumn{2}{c|}{\textbf{SIR$^2$(454)}}
& \multicolumn{2}{c}{\textbf{Avg}} \\
& PSNR$\uparrow$ & SSIM$\uparrow$ & PSNR$\uparrow$ & SSIM$\uparrow$ & PSNR$\uparrow$ & SSIM$\uparrow$ & PSNR$\uparrow$ & SSIM$\uparrow$ \\
\midrule
0.5  & 26.92 & 0.850 & 27.33 & 0.852 & 27.82 & 0.918 & 27.77 & 0.913 \\
\rowcolor{oursbg}
1.0 (default) & \textbf{27.14} & \textbf{0.853} & \textbf{27.35} & \textbf{0.853} & \textbf{27.95} & \textbf{0.918} & \textbf{27.90} & \textbf{0.913} \\
2.0  & 26.84 & 0.849 & 27.32 & 0.852 & 27.73 & 0.915 & 27.68 & 0.910 \\
\bottomrule
\end{tabular}
}
\end{table}

Table~\ref{tab:supp:lw_cycle} shows that the default $\lambda_{\text{cycle}} = 1$ achieves the best performance across all three benchmarks, while both halving and doubling the weight lead to modest degradation (within 0.3\,dB on Real and 0.2\,dB on SIR$^2$).
This confirms that performance is not critically sensitive to the exact choice of $\lambda_{\text{cycle}}$.

\begin{table}[h]
\centering
\caption{Sensitivity to Layer Contrastive Separation loss weight $\lambda_{\text{LCS}}$. All other weights are fixed at their default values.}
\label{tab:supp:lw_lcs}
\setlength{\tabcolsep}{5pt}
\resizebox{\linewidth}{!}{
\begin{tabular}{c|cc|cc|cc|cc}
\toprule
\multirow{2}{*}{$\lambda_{\text{LCS}}$}
& \multicolumn{2}{c|}{\textbf{Real(20)}}
& \multicolumn{2}{c|}{\textbf{Nature(20)}}
& \multicolumn{2}{c|}{\textbf{SIR$^2$(454)}}
& \multicolumn{2}{c}{\textbf{Avg}} \\
& PSNR$\uparrow$ & SSIM$\uparrow$ & PSNR$\uparrow$ & SSIM$\uparrow$ & PSNR$\uparrow$ & SSIM$\uparrow$ & PSNR$\uparrow$ & SSIM$\uparrow$ \\
\midrule
0.05 & 26.97 & 0.852 & 27.32 & 0.853 & 27.91 & 0.917 & 27.85 & 0.913 \\
\rowcolor{oursbg}
0.1 (default) & \textbf{27.14} & \textbf{0.853} & \textbf{27.35} & \textbf{0.853} & \textbf{27.95} & \textbf{0.918} & \textbf{27.90} & \textbf{0.913} \\
0.2  & 26.97 & 0.852 & 27.28 & 0.852 & 27.94 & 0.917 & 27.88 & 0.913 \\
\bottomrule
\end{tabular}
}
\end{table}

Table~\ref{tab:supp:lw_lcs} shows a similar trend for $\lambda_{\text{LCS}}$: the default value of 0.1 yields the best overall performance, while halving (0.05) or doubling (0.2) the weight causes negligible degradation ($\leq$0.17\,dB on Real, $\leq$0.01\,dB on SIR$^2$).
Both loss weights thus exhibit a broad plateau around their default values, confirming that PRISM's performance is robust to the exact hyperparameter choices.

\section{Stop-Gradient Ablation}
\label{sec:supp:sg}

\subsubsection{Motivation.}
The cycle consistency loss $\mathcal{L}_{\text{cycle}}$ applies stop-gradient ($\texttt{sg}$) to the first-pass predictions, preventing the cycle branch from back-propagating into the primary decomposition.
Without stop-gradient, the model could trivially minimize $\mathcal{L}_{\text{cycle}}$ by collapsing all first-pass predictions to a constant (mode collapse).
We empirically verify this claim by comparing training with and without stop-gradient.

\begin{table}[h]
\centering
\caption{Effect of removing stop-gradient from the cycle consistency loss.}
\label{tab:supp:sg}
\setlength{\tabcolsep}{5pt}
\resizebox{\linewidth}{!}{
\begin{tabular}{l|cc|cc|cc|cc}
\toprule
\multirow{2}{*}{Configuration}
& \multicolumn{2}{c|}{\textbf{Real(20)}}
& \multicolumn{2}{c|}{\textbf{Nature(20)}}
& \multicolumn{2}{c|}{\textbf{SIR$^2$(454)}}
& \multicolumn{2}{c}{\textbf{Avg}} \\
& PSNR$\uparrow$ & SSIM$\uparrow$ & PSNR$\uparrow$ & SSIM$\uparrow$ & PSNR$\uparrow$ & SSIM$\uparrow$ & PSNR$\uparrow$ & SSIM$\uparrow$ \\
\midrule
w/o stop-gradient & 26.82 & 0.849 & 27.23 & 0.853 & 27.84 & 0.916 & 27.78 & 0.911 \\
\rowcolor{oursbg}
w/ stop-gradient (default) & \textbf{27.14} & \textbf{0.853} & \textbf{27.35} & \textbf{0.853} & \textbf{27.95} & \textbf{0.918} & \textbf{27.90} & \textbf{0.913} \\
\bottomrule
\end{tabular}
}
\end{table}

\subsubsection{Analysis.}
Table~\ref{tab:supp:sg} confirms that removing stop-gradient consistently degrades performance across all benchmarks (--0.32\,dB on Real, --0.12\,dB on Nature, --0.11\,dB on SIR$^2$).
Without $\texttt{sg}$, the cycle branch can back-propagate into the first-pass predictions, allowing the model to reduce $\mathcal{L}_{\text{cycle}}$ without improving decomposition quality.
The stop-gradient operator prevents this shortcut and ensures that the cycle consistency signal provides a meaningful training signal.

\section{Timestep Sampling Strategy}
\label{sec:supp:timestep}

\subsubsection{Motivation.}
Standard flow matching~\cite{lipman2022flow} training samples $t \sim \mathcal{U}[0,1]$ and constructs interpolated inputs $x_t = (1-t)x_0 + tx_1$.
PRISM instead fixes $t=0$ and trains exclusively on the clean input latent $z_I$, since the velocity field $v^* = z_T - z_I$ is constant along the linear trajectory.
We verify that this simplification does not sacrifice performance by comparing against uniform $t$-sampling.
Both configurations are trained for the same number of iterations (50k) with identical hyperparameters, and we confirm that the uniform-$t$ variant has converged (training loss plateaus after $\sim$30k iterations).

\begin{table}[h]
\centering
\caption{Effect of timestep sampling strategy. PRISM uses $t\!=\!0$ (clean input) for both training and inference.}
\label{tab:supp:timestep}
\setlength{\tabcolsep}{5pt}
\resizebox{\linewidth}{!}{
\begin{tabular}{l|cc|cc|cc|cc}
\toprule
\multirow{2}{*}{$t$-sampling}
& \multicolumn{2}{c|}{\textbf{Real(20)}}
& \multicolumn{2}{c|}{\textbf{Nature(20)}}
& \multicolumn{2}{c|}{\textbf{SIR$^2$(454)}}
& \multicolumn{2}{c}{\textbf{Avg}} \\
& PSNR$\uparrow$ & SSIM$\uparrow$ & PSNR$\uparrow$ & SSIM$\uparrow$ & PSNR$\uparrow$ & SSIM$\uparrow$ & PSNR$\uparrow$ & SSIM$\uparrow$ \\
\midrule
$t \sim \mathcal{U}[0,1]$ & 22.99 & 0.775 & 24.08 & 0.812 & 25.47 & 0.885 & 25.27 & 0.876 \\
\rowcolor{oursbg}
$t = 0$ (default) & \textbf{27.14} & \textbf{0.853} & \textbf{27.35} & \textbf{0.853} & \textbf{27.95} & \textbf{0.918} & \textbf{27.90} & \textbf{0.913} \\
\bottomrule
\end{tabular}
}
\end{table}

Table~\ref{tab:supp:timestep} reveals a substantial performance gap between uniform $t$-sampling and the fixed $t\!=\!0$ strategy ($-$4.15\,dB on Real, $-$2.48\,dB on SIR$^2$).
Although the optimal velocity $v^* = z_T - z_I$ is theoretically constant along the linear trajectory, uniform $t$-sampling trains the model on interpolated inputs $z_t = (1\!-\!t)z_I + tz_T$ that are never encountered at inference, where the model always receives the clean input $z_I$.
Fixing $t\!=\!0$ eliminates this train-test mismatch, allowing the network to focus its capacity entirely on the operationally relevant input distribution.

\section{Discussion of Design Choices}
\label{sec:supp:discussion}

\subsubsection{Why does the trivial solution not arise?}
\label{sec:supp:trivial}
A natural concern with the latent decomposition $z_I \approx z_T + z_R$ is whether the model can exploit a trivial solution:
predict $v_\theta(z_I) = 0$, yielding $\hat{z}_T = z_I$ and $\hat{z}_R = 0$.
This would collapse the decomposition to $\hat{T} = I$ (transmission equals the input) and $\hat{R} = \mathcal{D}(\mathbf{0})$ (a gray image), effectively ignoring the reflection layer entirely.
PRISM prevents this collapse through its pixel-space transmission losses.
The training objective includes $\|\hat{T} - T_\text{gt}\|_1$ and $\mathcal{L}_\text{LPIPS}(\hat{T}, T_\text{gt})$, which directly penalize the decoded transmission $\hat{T} = \mathcal{D}(\hat{z}_T)$ against the ground-truth transmission.
If the model predicts $v_\theta = 0$, then $\hat{T} = \mathcal{D}(z_I) = I \neq T_\text{gt}$ because the input mixture contains reflections.
These pixel-space losses therefore impose a non-zero gradient on $v_\theta$ whenever $I \neq T_\text{gt}$ (assuming the pretrained VAE decoder has a non-degenerate Jacobian, which holds in practice), forcing the model to learn a meaningful decomposition.
Note that PRISM does \emph{not} apply any pixel-space loss on the reflection layer $\hat{R}$;
the reflection is recovered solely via the flow matching target $\hat{z}_R = -v_\theta$ and the LCC/LCS objectives.
As shown in Sec.~\ref{sec:supp:reflection}, this leads to a systematic brightness bias in the decoded reflections,
which can be corrected by a simple post-hoc affine calibration.
The key insight is that the pixel-space transmission loss alone is sufficient to prevent the trivial solution,
while the latent-space objectives (LCC, LCS) further regularize the decomposition to produce structurally accurate reflections.

\subsubsection{Is LCC sensitive to batch size?}
\label{sec:supp:batchsize}
The LCC loss constructs cross-image compositions by cyclically swapping reflection latents within a mini-batch.
With the default batch size of 2, only a single swap pair is formed per step, raising the question of whether a larger swap pool would improve training.
However, LCC is a \emph{reconstruction consistency} loss — it verifies that the model can correctly decompose a novel composition — rather than a contrastive objective that benefits from a large number of negatives.
The diversity of compositions encountered during training is determined by the dataset itself (each step samples a fresh image pair), not by the number of simultaneous swaps within a single batch.
Furthermore, the LCS loss, which does use InfoNCE~\cite{oord2018representation}, operates on channels within a single image (transmission vs.\ reflection features) rather than across batch samples, so it is similarly unaffected by batch size.
We therefore expect performance to be largely insensitive to batch size beyond $N\!=\!2$.

\section{Additional Qualitative Results}
\label{sec:supp:qualitative}

\subsubsection{Overview.}
We provide additional visual comparisons on Real~\cite{zhang2018single}, Nature~\cite{li2020single}, SIR$^2$~\cite{wan2017benchmarking} (Object, Postcard, Wild), and the in-the-wild OpenRR 1K~\cite{yang2025openrr} benchmark.
For each benchmark, we compare PRISM against representative baselines along with the input mixture.
All images are shown at their original resolution; crops are used where fine-grained detail is relevant.

\subsubsection{Real and Nature.}
\begin{figure}[h]
    \centering
    \includegraphics[width=\linewidth]{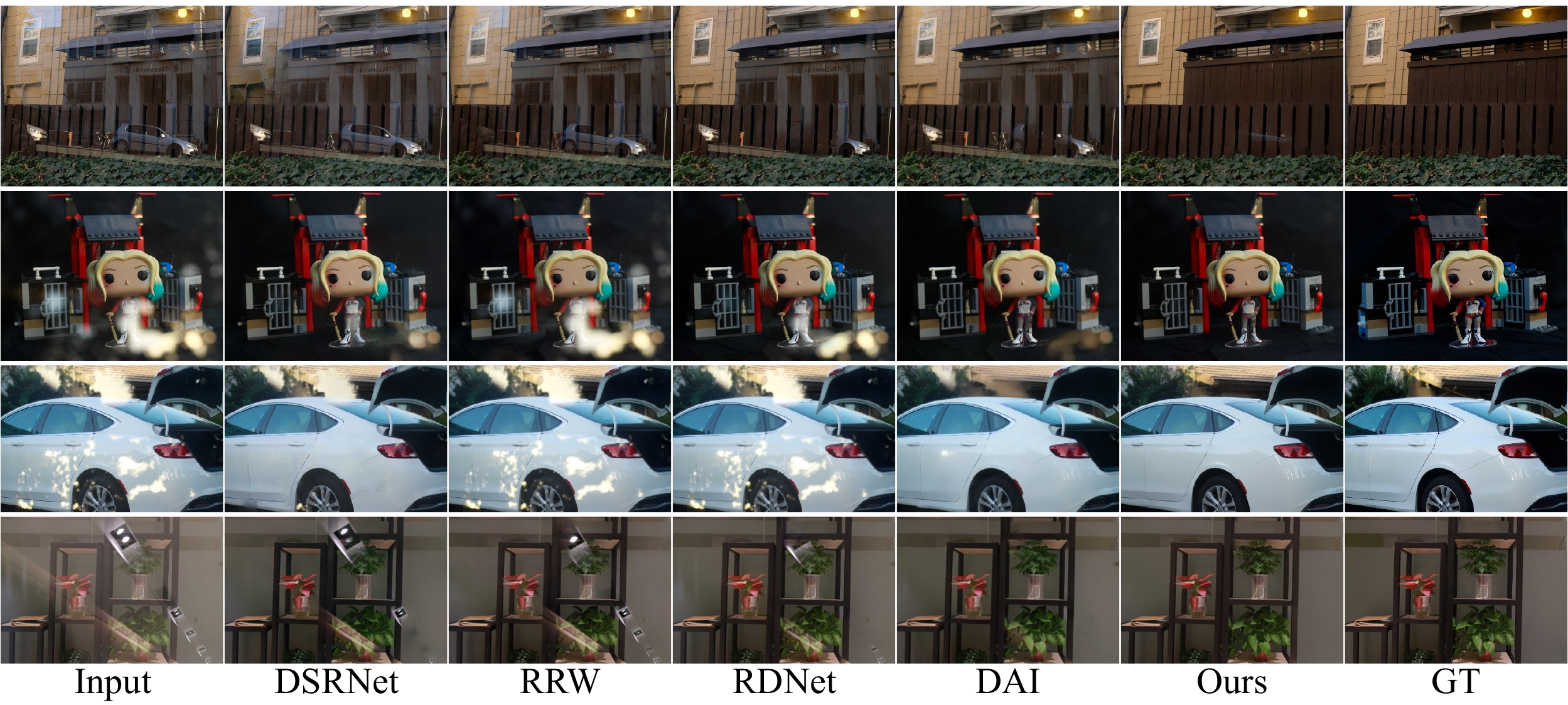}
    \caption{Additional qualitative comparisons on the Real~\cite{zhang2018single} (top three rows) and Nature~\cite{li2020single} (bottom row) benchmarks.}
    \label{fig:supp:qual:real}
\end{figure}

Fig.~\ref{fig:supp:qual:real} shows additional results on the Real~\cite{zhang2018single} and Nature~\cite{li2020single} benchmarks.
PRISM recovers sharper transmission layers with fewer residual reflection artifacts, while competing methods leave residual artifacts or introduce color shifts.

\subsubsection{SIR$^2$ (Object, Postcard, Wild).}
\begin{figure}[h]
    \centering
    \includegraphics[width=\linewidth]{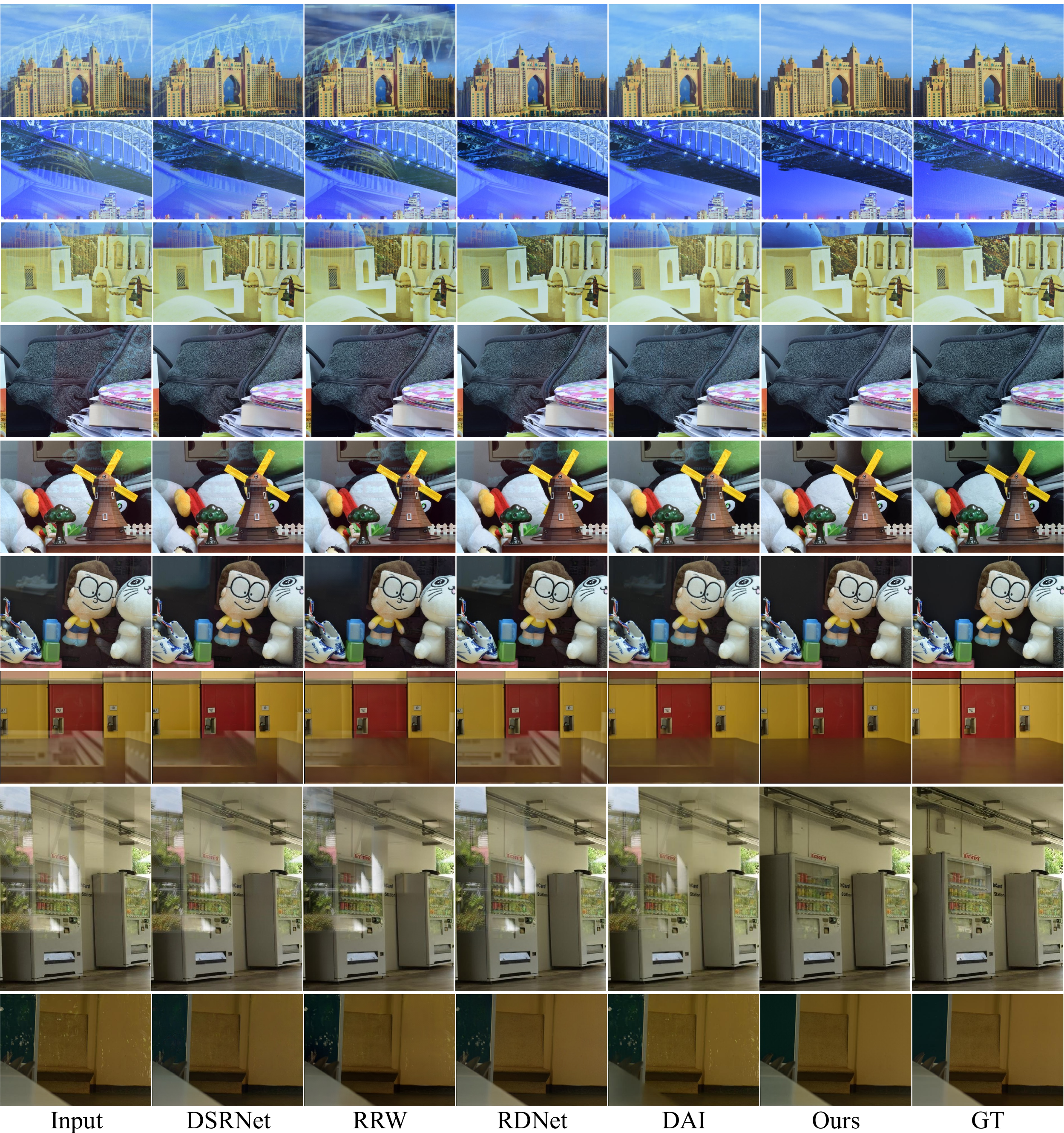}
    \caption{Additional qualitative comparisons on the SIR$^2$~\cite{wan2017benchmarking} benchmark.
    From top to bottom: Postcard (rows 1--3), Object (rows 4--6), and Wild (rows 7--9).}
    \label{fig:supp:qual:sir2}
\end{figure}

Fig.~\ref{fig:supp:qual:sir2} presents results across all three SIR$^2$~\cite{wan2017benchmarking} subsets.
PRISM produces cleaner separations across diverse reflection types, including structured text reflections (Postcard), specular highlights (Object), and complex outdoor scenes (Wild).

\subsubsection{OpenRR 1K (In-the-wild).}
\begin{figure}[h]
    \centering
    \includegraphics[width=\linewidth]{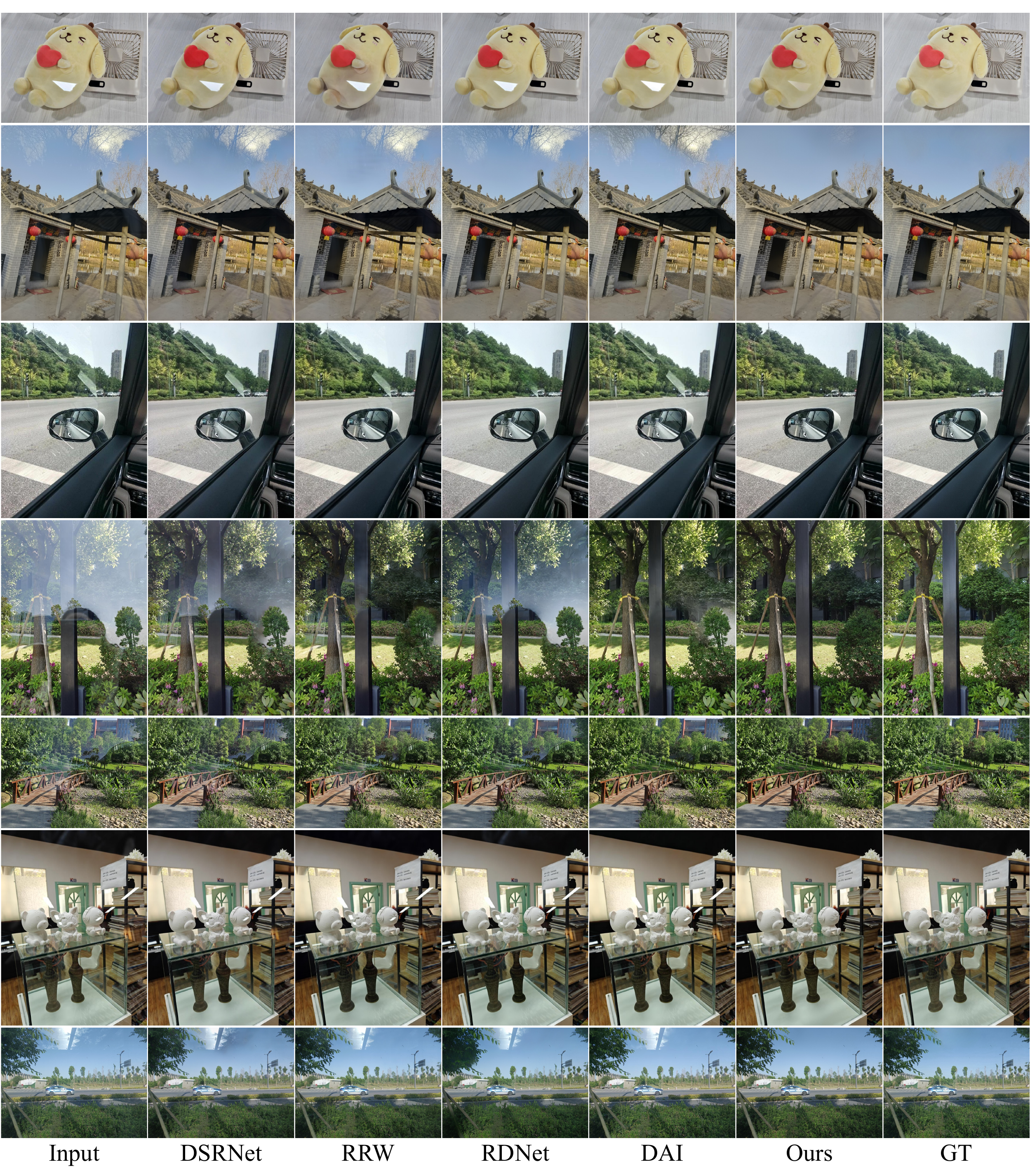}
    \caption{Additional qualitative comparisons on the OpenRR 1K test set~\cite{yang2025openrr}.
    All methods are evaluated without training on this dataset.}
    \label{fig:supp:qual:openrr}
\end{figure}

Fig.~\ref{fig:supp:qual:openrr} shows results on the OpenRR 1K~\cite{yang2025openrr} test set, where all methods are evaluated without training on this dataset.
PRISM demonstrates stronger generalization to diverse in-the-wild reflection scenarios, producing cleaner results with fewer residual artifacts.

\section{Generative Prior Comparison}
\label{sec:supp:generative}

\begin{figure}[h]
    \centering
    \includegraphics[width=\linewidth]{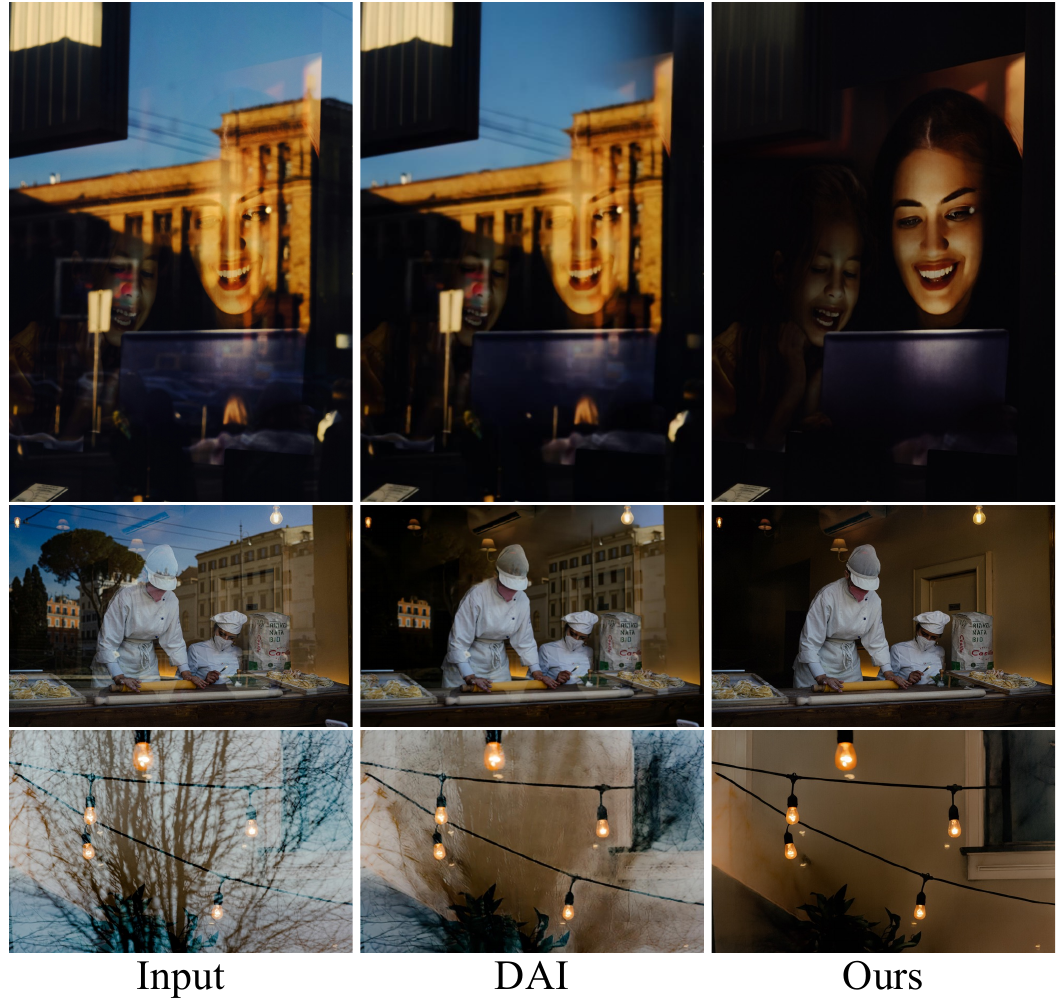}
    \caption{Visual comparison of generative-prior-based reflection removal methods on real-world images from Unsplash.
    From left to right: input $I$, DAI~\cite{hu2025dereflection}, and Ours.}
    \label{fig:supp:generative}
\end{figure}

Among the compared methods, both DAI~\cite{hu2025dereflection} and PRISM leverage pretrained generative models for reflection removal:
DAI builds on a one-step diffusion framework with ControlNet conditioning,
while PRISM reformulates the task as latent flow matching using FLUX~\cite{blackforest2024flux}.
To compare how these two generative-prior-based approaches handle real-world reflections beyond existing benchmarks,
we evaluate them on challenging images sourced from Unsplash\footnote{\url{https://unsplash.com}}, which exhibit diverse and intense reflection conditions not represented in standard evaluation sets such as SIR$^2$ or OpenRR.
Fig.~\ref{fig:supp:generative} presents the visual comparison.

\section{Pixel-Only and Pixel-Cycle Ablations}
\label{sec:supp:pixelablation}

To further isolate the contribution of latent-space supervision, we conduct two additional ablations.

\begin{table}[h]
\centering
\caption{Pixel-only and pixel-cycle ablations. ``$\mathcal{L}_\text{pixel}$ only'' removes all latent-space losses. ``Pixel-cycle'' moves the cycle consistency to pixel space. VAE/step counts the number of VAE encoder/decoder calls per training step.}
\label{tab:supp:pixelablation}
\setlength{\tabcolsep}{4pt}
\resizebox{\linewidth}{!}{
\begin{tabular}{l|cc|cc|cc|c}
\toprule
\multirow{2}{*}{Setting}
& \multicolumn{2}{c|}{\textbf{Real(20)}}
& \multicolumn{2}{c|}{\textbf{Nature(20)}}
& \multicolumn{2}{c|}{\textbf{SIR$^2$(454)}}
& VAE/step \\
& PSNR$\uparrow$ & SSIM$\uparrow$ & PSNR$\uparrow$ & SSIM$\uparrow$ & PSNR$\uparrow$ & SSIM$\uparrow$ & \\
\midrule
$\mathcal{L}_\text{pixel}$ only          & 26.25 & — & 27.28 & — & 27.65 & — & 2$\times$ \\
Pixel-cycle                               & 26.93 & — & 27.31 & — & 27.94 & — & 5$\times$ \\
\rowcolor{oursbg}
PRISM (full)                              & \textbf{27.14} & \textbf{0.853} & \textbf{27.35} & \textbf{0.853} & \textbf{27.95} & \textbf{0.918} & 2$\times$ \\
\bottomrule
\end{tabular}}
\end{table}

\subsubsection{Analysis.}
Training with $\mathcal{L}_\text{pixel}$ only (no latent loss) drops Real by $0.89$\,dB (Tab.~\ref{tab:supp:pixelablation}), confirming that latent-space supervision matters beyond pixel reconstruction.
Moving the cycle consistency to pixel space drops Real by $0.21$\,dB while costing $2.5{\times}$ more VAE forward passes per training step (2 extra decodes + 1 encode to compute cycle consistency in pixel space).
The latent cycle thus achieves better quality at lower training cost, because cycle consistency in latent space avoids the information bottleneck of decoding and re-encoding.

\section{Failure Case Analysis}
\label{sec:supp:failure}

\subsubsection{Overview.}
While PRISM achieves strong performance across diverse benchmarks, there exist challenging conditions where all current methods, including PRISM, produce suboptimal results.
We present an additional failure mode that remains an open challenge in single-image reflection removal (the dense/saturated reflection case is discussed in the main paper).
Fig.~\ref{fig:supp:failure2} compares PRISM with RDNet~\cite{zhao2025reversible} and DAI~\cite{hu2025dereflection}.

\subsubsection{Structural artifacts from the glass surface.}
\begin{figure}[h]
    \centering
    \includegraphics[width=\linewidth]{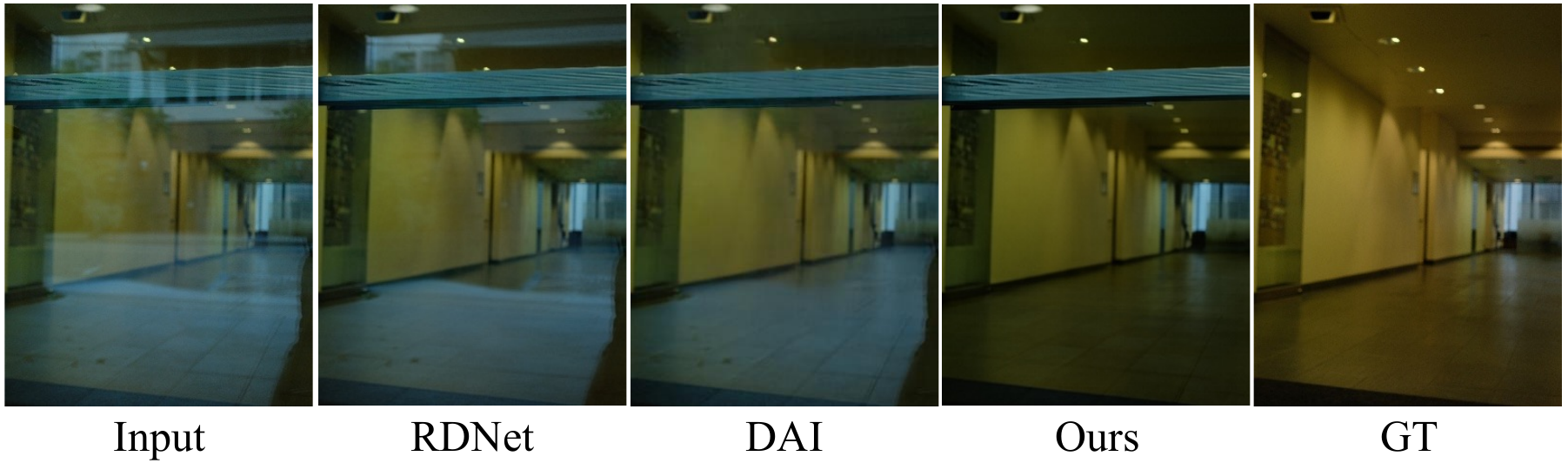}
    \caption{Failure case: structural glass artifacts on SIR$^2$~\cite{wan2017benchmarking}.
    From left to right: input $I$, RDNet~\cite{zhao2025reversible}, DAI~\cite{hu2025dereflection}, Ours, and ground truth $T$.}
    \label{fig:supp:failure2}
\end{figure}

In some SIR$^2$ samples, the physical structure of the glass panel used during data capture (\eg, frames, support beams, or canopy elements) appears as strong, sharp-edged patterns superimposed on the scene.
These structural artifacts differ from typical scene reflections in that they exhibit hard geometric edges rather than soft, blended intensity variations.
All compared methods struggle to remove these patterns, as shown in Fig.~\ref{fig:supp:failure2}, because the sharp edges are difficult to distinguish from genuine transmission content.

\subsubsection{Discussion.}
This limitation is shared across all compared methods and stems from the fundamental assumption that scene content consists of smooth intensity variations.
Structural glass artifacts introduce sharp geometric patterns that are difficult to distinguish from genuine transmission content.
Future work may benefit from edge-aware separation mechanisms that can distinguish physical glass structures from scene edges.

\end{document}